\documentclass[10pt,twocolumn,letterpaper]{article}

\usepackage[pagenumbers]{cvpr}

\usepackage{multicol}
\usepackage{multirow}
\usepackage{marvosym}
\usepackage{titletoc}
\usepackage{pifont}

\usepackage[table,dvipsnames]{xcolor}

\newcommand{\cmark}{\ding{51}}
\newcommand{\xmark}{\ding{55}}

\definecolor{gen_blue}{RGB}{99,113,250}
\definecolor{gen_red}{RGB}{239,99,75}
\definecolor{gen_green}{RGB}{0,180,139}
\definecolor{gen_gray}{RGB}{165,165,165}

\definecolor{cvprblue}{rgb}{0.21,0.49,0.74}
\usepackage[pagebackref,breaklinks,colorlinks,allcolors=cvprblue]{hyperref}

\usepackage[accsupp]{axessibility}

\newcommand\blfootnote[1]{%
\begingroup
\renewcommand\thefootnote{}{}\footnote{#1}%
\addtocounter{footnote}{-1}%
\endgroup
}

\def\paperID{756}
\def\confName{CVPR}
\def\confYear{2026}

\title{\vspace{-0.4cm}\includegraphics[width=0.06\linewidth]{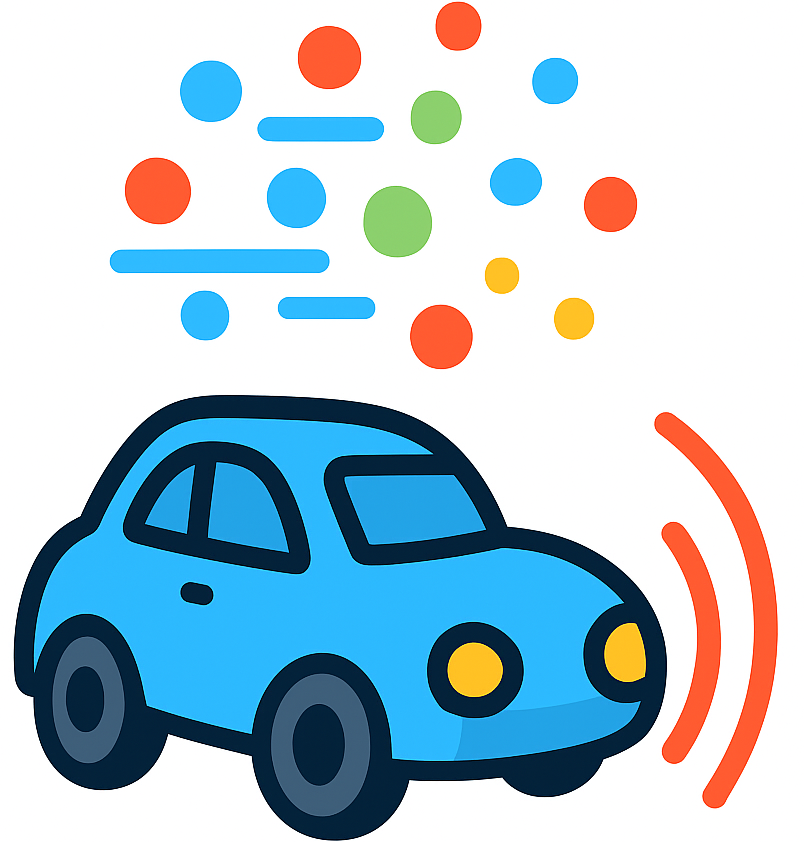}~U4D: Uncertainty-Aware 4D World Modeling from LiDAR Sequences}

\author{
Xiang Xu$^{1}$ \quad Alan Liang$^{2}$ \quad Youquan Liu$^{3}$ \quad Linfeng Li$^{2}$ \quad Lingdong Kong$^{2,\dagger}$
\\
Ziwei Liu$^{4}$ \quad Qingshan Liu$^{5,6,\textrm{\Letter}}$
\\[0.5ex]
{\small$^1$Nanjing University of Aeronautics and Astronautics \quad $^2$National University of Singapore \quad $^3$Fudan University}
\\
{\small$^4$S-Lab, Nanyang Technological University \quad $^5$Nanjing University of Posts and Telecommunications \quad $^6$SKL-TI}
}

\begin{document}

\twocolumn[{
    \renewcommand\twocolumn[1][]{#1}
    \maketitle
    \begin{center}
    \centering
    \captionsetup{type=figure}
    \vspace{-0.5cm}
    \includegraphics[width=\linewidth]{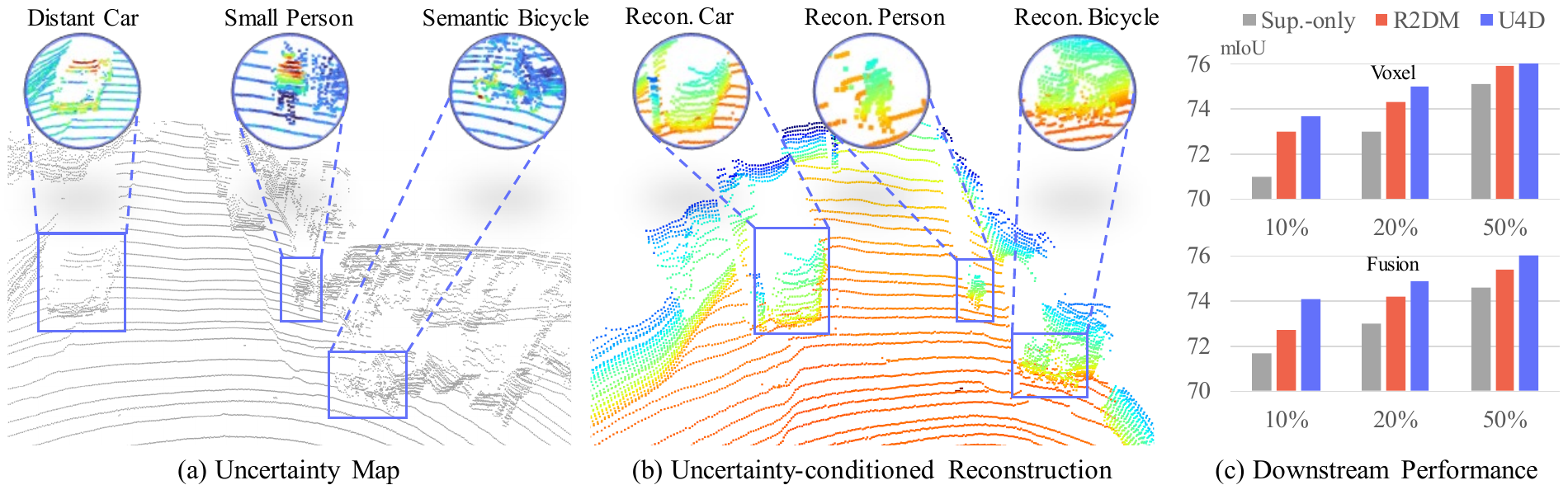}
    \vspace{-0.6cm}
    \caption{\textbf{Illustration of the proposed \textbf{U4D} framework} for uncertainty-aware LiDAR scene generation. \textbf{(a)} U4D first estimates the spatial uncertainty maps, highlighting regions that are challenging for perception, such as distant or partially occluded objects, small-scale instances, and semantically ambiguous areas. \textbf{(b)} Conditioned on these uncertainty regions, U4D performs scene completion in a ``hard-to-easy'' manner, progressively reconstructing the entire scene with enhanced fidelity in uncertain regions. \textbf{(c)} The generated uncertainty-aware scenes can further benefit downstream perception tasks by improving robustness and recognition performance.}
    \label{fig:teaser}
    \vspace{0.05cm}
    \end{center}
}]
\blfootnote{($\dagger$) Project lead. $(\textrm{\Letter})$ Corresponding author.}


\begin{abstract}
Modeling dynamic 3D environments from LiDAR sequences is central to building reliable 4D worlds for autonomous driving and embodied AI. Existing generative frameworks, however, often treat all spatial regions uniformly, overlooking the varying uncertainty across real-world scenes. This uniform generation leads to artifacts in complex or ambiguous regions, limiting realism and temporal stability. In this work, we present \textbf{U4D}, an uncertainty-aware framework for 4D LiDAR world modeling. Our approach first estimates spatial uncertainty maps from a pretrained segmentation model to localize semantically challenging regions. It then performs generation in a ``hard-to-easy" manner through two sequential stages: (1) \textit{uncertainty-region modeling}, which reconstructs high-entropy regions with fine geometric fidelity, and (2) \textit{uncertainty-conditioned completion}, which synthesizes the remaining areas under learned structural priors. To further ensure temporal coherence, U4D incorporates a mixture of spatio-temporal (MoST) block that adaptively fuses spatial and temporal representations during diffusion. Extensive experiments show that U4D produces geometrically faithful and temporally consistent LiDAR sequences, advancing the reliability of 4D world modeling for autonomous perception and simulation.
\end{abstract}

\vspace{-0.4cm}
\section{Introduction}
\label{sec:introduction}

Modeling dynamic 3D environments from LiDAR sequences is fundamental to constructing reliable 4D world models that enable autonomous systems to perceive, simulate, and interact with the physical world over time~\cite{kong2025survey,wen2025survey,munoz2021survey}. LiDAR provides precise geometric and depth information, forming the basis for perception, mapping, and planning in autonomous driving~\cite{li2021deep,li2024place3d,xu2025frnet}, robotics~\cite{guo2021deep,qin2020lins,liu2021efficient}, and 3D scene reconstruction~\cite{zhang2014loam,wolcott2014visual}. However, collecting large-scale, diverse, and annotated LiDAR data remains costly and labor-intensive~\cite{xu2024superflow,gao2022we,sun2024empirical,lyu2025alise}, motivating increasing interest in generative LiDAR modeling for scalable simulation, data augmentation, and pretraining.

Recent advances have explored LiDAR scene generation via adversarial, variational, and diffusion-based generative frameworks~\cite{wen2025survey,caccia2019deep,li2025scaling}. Early efforts focus on object-level point clouds~\cite{achlioptas2018learning,gao2019sdm-net,kim2021setvae,mo2019structurenet}, whereas recent methods such as LiDARGen~\cite{zyrianov2022lidargen}, LiDM~\cite{ran2024lidm}, and LiDARCrafter~\cite{liang2026lidarcrafter} synthesize large-scale and even dynamic LiDAR scenes. Despite these advances, existing methods treat spatial regions equally during generation, ignoring the varying semantic difficulty of real-world data. This uniform assumption often causes unreliable reconstruction in geometrically or semantically complex regions, such as thin poles, moving objects, and distant surfaces, where predictive confidence is low.

We observe that reliable LiDAR world modeling requires understanding the underlying uncertainty itself. As shown in \cref{fig:teaser}~(a), real LiDAR observations exhibit inherently non-uniform difficulty: while some areas are well-defined, others -- such as occluded areas, small-scale structures, or semantically ambiguous regions -- remain uncertain. Ignoring this asymmetry leads to geometric artifacts and temporal instability. Inspired by how humans resolve ambiguous regions before perceiving global context, we propose to model uncertainty explicitly, generating difficult regions first as structural anchors for the rest of the scene.

To this end, we propose \textbf{U4D}, an uncertainty-aware framework for 4D LiDAR world modeling. U4D leverages spatial uncertainty as a structural prior to guide scene generation. Our framework first estimates an uncertainty map from a pretrained LiDAR segmentation network to localize semantically ambiguous or geometrically unreliable regions. It then performs two sequential stages of generation: (1) an \textit{uncertainty-region diffusion} stage, which focuses on reconstructing high-entropy regions with fine geometric fidelity, and (2) an \textit{uncertainty-conditioned completion} stage, which synthesizes the remaining areas conditioned on these reconstructed structures (\cref{fig:teaser}~(b)). The two stages share a unified latent scene representation, enabling global contextual cues to refine local uncertainty and ensuring geometric and temporal consistency across the generated 4D scenes.

To further ensure stable temporal evolution, \textbf{U4D} integrates a \textbf{M}ixture \textbf{o}f \textbf{S}patio-\textbf{T}emporal (\textbf{MoST}) block, which explicitly decomposes and adaptively fuses spatial and temporal representations within the diffusion process. This design enables the generation of LiDAR sequences that are both geometrically precise and temporally coherent. Extensive experiments on the \textit{nuScenes}~\cite{caesar2020nuscenes} and \textit{SemanticKITTI}~\cite{behley2019semantickitti} datasets demonstrate that U4D consistently outperforms existing LiDAR generation frameworks in terms of geometric fidelity, temporal stability, and downstream generalization (\cref{fig:teaser}~(c)).

The main contributions of this work are as follows:
\begin{itemize}
    \item We introduce the first uncertainty-aware LiDAR generation framework that explicitly models spatial difficulty to enhance reliability in 4D world modeling.
    \item We design a two-stage hard-to-easy generation paradigm that reconstructs uncertain regions first and then completes the full scene under these priors. 
    \item We develop a Mixture of Spatio-Temporal (MoST) block that ensures temporal consistency across frames by adaptively balancing spatial geometry and temporal dynamics.
\end{itemize}

\section{Related Work}
\label{sec:related_work}

\noindent\textbf{LiDAR Scene Understanding.}
LiDAR sensors provide accurate geometric measurements of 3D environments~\cite{hong20224dDSNet,li2025_3eed,xu2025visual,jaritz2020xMUDA,liang2025pi3det,xie2025benchmarking}. Due to the sparse and irregular nature of LiDAR point clouds, existing methods typically transform LiDAR data into alternative representations for efficient processing, including raw points~\cite{qi2017pointnet,qi2017pointnet++,hu2020randla,shuai2021baflac}, bird's-eye-view~\cite{zhang2020polarnet,zhou2021panoptic-polarnet,chen2021polarstream}, range images~\cite{ando2023rangevit,kong2023rangeformer,xu2025frnet,xu2020squeezesegv3}, sparse voxels~\cite{choy2019minkunet,zhu2021cylinder3d,yan2018second,hong2021dsnet,yin2021centerpoint}, and multi-view fusion~\cite{tang2020spvcnn,xu2021rpvnet,liong2020amvnet,zhuang2021pmf,liu2023uniseg}. While achieving strong performance across tasks such as segmentation and detection, these methods heavily rely on large-scale annotated datasets, which are costly to acquire~\cite{wang2026forging,chen2023clip2Scene,chen2023towards,li2025seeground}. To alleviate this dependency, recent works explore semi-supervised~\cite{kong2025lasermix++,li2023lim3d,wu2024patchteacher}, weakly-supervised~\cite{liu2022box2seg,meng2022towards,xu2020weakly}, and self-supervised~\cite{xu2025lima,sautier2022slidr,puy2024scalr,yu2022point-bert,xu2026superflow++,kong2026largead} learning strategies, aiming to maintain high performance while reducing the need for extensive labeled data.

\noindent\textbf{LiDAR Scene Generation.}
Generative models learn distributions of 3D points and synthesize realistic scenes. Early GAN-based~\cite{achlioptas2018learning,shu2019treegcn,valsesia2018learning,caccia2019deep} and VAE-based~\cite{gao2021tm-net,gao2019sdm-net,kim2021setvae,litany2018deformable,mo2019structurenet,klokov2020discrete} methods focus on object-level synthesis but fail to scale to outdoor scenes. LiDARGen~\cite{zyrianov2022lidargen} pioneers large-scale LiDAR generation via range-image projection and score-based modeling, while R2DM~\cite{nakashima2024r2dm} and R2Flow~\cite{nakashima2025r2flow} adopt diffusion models~\cite{ho2020ddpm} for faithful synthesis. Subsequent works~\cite{ran2024lidm,wu2024text2lidar,hu2024rangeldm,wu2025weathergen,liu2026lalalidar,liu2026veila,zhu2025spiral,zyrianov2025lidardm} enhance controllability through conditional diffusion. UltraLiDAR~\cite{xiong2023ultralidar} integrates voxelized LiDAR in a BEV-based VQ-VAE~\cite{van2017vqvae} framework for spatial compactness. For temporal modeling, ViDAR~\cite{yang2024vidar} predicts future LiDAR frames from historical images, while LiDARCrafter~\cite{liang2026lidarcrafter} employs a two-stage autoregressive framework. These methods treat spatial regions uniformly, overlooking varying semantic difficulty and uncertainty. In contrast, our work explicitly estimates and models spatial uncertainty, guiding generation in a ``hard-to-easy'' manner for improved structural fidelity.

\begin{figure*}[t]
    \centering
    \includegraphics[width=\linewidth]{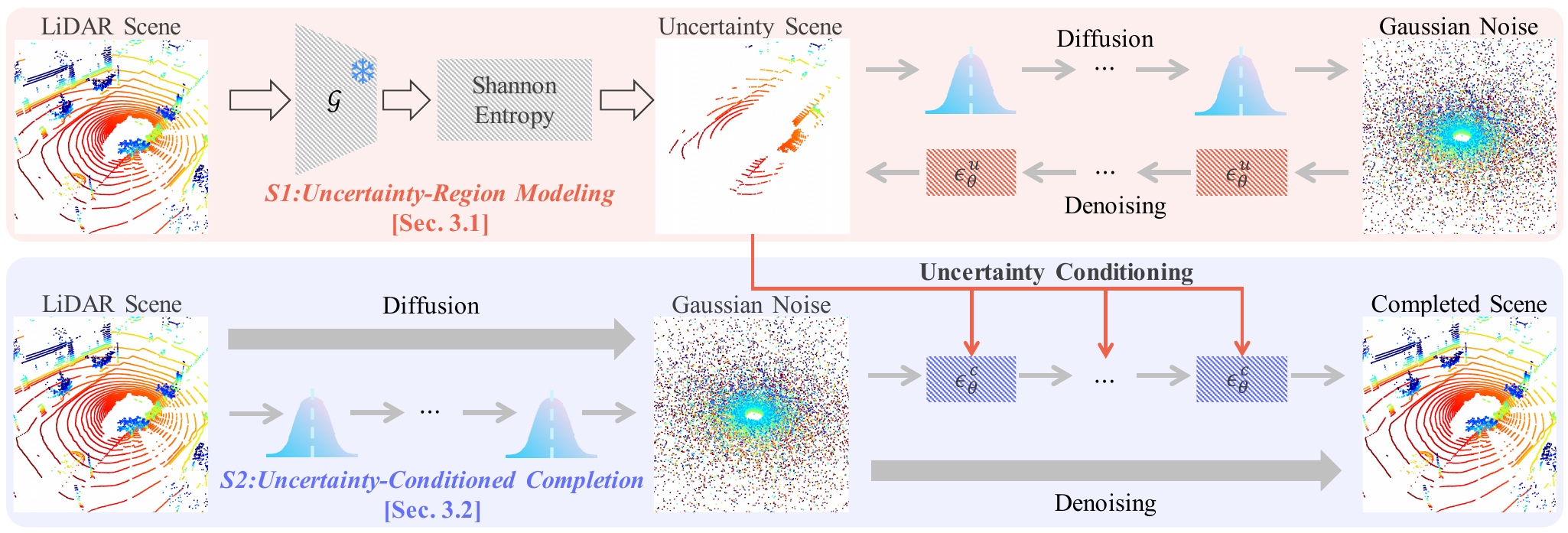}
    \vspace{-0.6cm}
    \caption{\textbf{Overview of the U4D framework.} U4D generates LiDAR scenes in a ``hard-to-easy'' manner through two stages. (1) It first estimates spatial uncertainty using a pretrained segmentation model $\mathcal{G}$ based on Shannon Entropy, and performs an unconditional diffusion process to reconstruct high-fidelity geometry within the uncertain regions (\cf~\cref{subsec:uncertainty_gen}). (2) It then conducts uncertainty-conditioned completion, synthesizing the remaining scene areas guided by the reconstructed structures to ensure global consistency (\cf~\cref{subsec:scene_completion}).}
    \label{fig:framework}
    \vspace{-0.4cm}
\end{figure*}

\noindent\textbf{Uncertainty Modeling in 3D.}
Uncertainty modeling is critical for reliable perception in safety-critical applications such as autonomous driving~\cite{ashukha2020pitfalls,hendrycks2016baseline,survey_vla4ad}. In the image domain, prior works improve model robustness through deterministic confidence estimation~\cite{malinin2018predictive,sensoy2018evidential}, Bayesian inference~\cite{denker1990transforming,gal2016dropout,hernandez2016black,kendall2017uncertainties}, ensemble learning~\cite{gustafsson2020evaluating,lakshminarayanan2017simple,rahaman2021uncertainty}, test-time augmentations~\cite{ayhan2018test,lyzhov2020greedy}, and post-hoc calibration~\cite{guo2017calibration,oksuz2023towards,ding2021local,kassapis2021calibrated}. In the 3D domain, Zhang~\etal~\cite{zhang2021point} propose probabilistic point embeddings to enhance uncertainty estimation, while SalsaNext~\cite{cortinhal2020salsanext} integrates Bayesian inference to capture epistemic and aleatoric uncertainty in LiDAR segmentation. Calib3D~\cite{kong2025calib3d} further introduces depth-aware scaling to calibrate confidence across spatial distributions. In contrast, our work is the first to incorporate uncertainty modeling into a generative framework, explicitly leveraging estimated uncertainty regions as structural priors to guide scene synthesis and improve downstream reliability.

\section{Methodology}
\label{sec:methodology}

We propose \textbf{U4D}, the first uncertainty-aware generative framework for LiDAR scene synthesis. As illustrated in \cref{fig:framework}, U4D generates LiDAR scenes in a ``hard-to-easy'' manner through two sequential stages. (1) It estimates an uncertainty map from a real scan using a pretrained segmentation model and employs an unconditional diffusion process to reconstruct high-fidelity uncertain regions (\cf~\cref{subsec:uncertainty_gen}). (2) Conditioned on these reconstructed areas, U4D completes the remaining scene to ensure structural integrity and global coherence (\cf~\cref{subsec:scene_completion}). Both stages share latent representations, enabling global context to refine local uncertainty. To maintain temporal stability, U4D integrates a \textbf{M}ixture \textbf{o}f \textbf{S}patio-\textbf{T}emporal (\textbf{MoST}) block within the diffusion backbone. The MoST block decouples and adaptively fuses spatial and temporal features, enabling the generation of LiDAR sequences that are both geometrically precise and temporally coherent (\cf~\cref{subsec:most}).

\subsection{Uncertainty Measurement in 3D}
\label{subsec:uncertainty_gen}

\noindent
Real-world LiDAR scenes exhibit non-uniform difficulty across spatial regions. Some structures, such as ground or buildings, are geometrically stable and semantically consistent, while others are inherently uncertain due to factors such as \textit{distance-induced sparsity}, \textit{occlusion}, \textit{small-scale objects}, or \textit{semantic ambiguity} between visually similar categories. These uncertainty-prone regions frequently appear at long ranges, around object boundaries, or in areas of low point density, leading to inconsistent predictions in safety-critical perception tasks. Explicitly identifying and modeling these regions allows generative models to focus first on structurally unstable and perceptually ambiguous areas before extending to the entire scene, thereby producing more realistic priors and improving downstream robustness.

\noindent\textbf{Uncertainty Maps.}
To localize uncertainty regions, we derive a per-point uncertainty map from a pretrained LiDAR semantic segmentation model. Given a point cloud $\mathcal{P}$ and a segmentation model $\mathcal{G}$, the model outputs per-point semantic logits $\mathbf{Z} = \mathcal{G}(\mathcal{P}) \in \mathbb{R}^{|\mathcal{P}| \times C}$, where $C$ denotes the number of classes. After softmax normalization, we obtain the class probability $D_{c}(\mathbf{p})$ for each point $\mathbf{p} \in \mathcal{P}$. The semantic uncertainty is quantified via Shannon Entropy~\cite{shannon1948entropy, kendall2017uncertainties}:
\begin{equation}
    H(\mathbf{p}) = -\sum_{c=1}^{C} D_{c}(\mathbf{p}) \log D_{c}(\mathbf{p})~.
\end{equation}
Higher entropy values correspond to greater ambiguity, typically found near class boundaries or within distant, partially occluded areas where the LiDAR returns become sparse. To maintain a consistent proportion of uncertain points across scenes, we select the top-$K$ points with the highest entropy to form a sparse uncertainty point cloud $\mathcal{P}^{u}$, which highlights geometrically or semantically uncertain regions. This point cloud serves as a structural prior for subsequent uncertainty-aware generation.

\noindent\textbf{Uncertainty-Region Representation.}
To efficiently represent and model these uncertainty regions, we transform $\mathcal{P}^{u}$ into a range-view format, following prior works~\cite{nakashima2024r2dm,liu2026lalalidar}. Each frame is projected onto a range image $\mathbf{x}_{0}^{u} \in \mathbb{R}^{H \times W \times 2}$, where each pixel encodes normalized depth and reflectance values within $[-1, 1]$. A binary mask $\mathbf{m}^{u} \in \{0, 1\}^{H \times W \times 1}$ simultaneously indicates valid LiDAR returns in the 2D projection. This representation preserves the geometric continuity of uncertainty regions while providing dense spatial layout suitable for generative modeling.

\noindent\textbf{Uncertainty Diffusion Modeling.}
To model the generative distribution of uncertainty regions, we employ an unconditional diffusion model $\epsilon_{\theta}^{u}$ trained on the range-view representation. Following the standard denoising diffusion formulation~\cite{ho2020ddpm}, Gaussian noise is progressively added to the clean data $\mathbf{x}_{0}^{u}$ with a predefined variance schedule $\{\beta_t\}_{t=1}^{T}$:
\begin{equation}
    \label{eq:forward}
    q(\mathbf{x}_{t}^{u} \mid \mathbf{x}_{0}^{u}) = \mathcal{N}(\mathbf{x}_{t}^{u}; \sqrt{\bar{\alpha}_{t}} \mathbf{x}_{0}^{u}, (1 - \bar{\alpha}_{t}) \mathbf{I})~,
\end{equation}
where $\bar{\alpha}_{t} = \prod_{s=1}^{t}(1 - \beta_{s})$. During training, $\epsilon_{\theta}^{u}$ learns to predict the injected Gaussian noise $\epsilon^{u}$ while reconstructing the binary occupancy mask $\mathbf{m}^{p}$ to preserve spatial validity. The training objective combines the diffusion noise prediction loss and an auxiliary mask supervision:
\begin{equation}
    \mathcal{L}_{u} = \mathbb{E}_{t, \mathbf{x}_{0}^{u}, \epsilon^{u}} \left[\| \epsilon^{u} - \epsilon_{\theta}^{u}(\mathbf{x}_{t}^{u}, t) \|_{2}^{2} \right] + \lambda \mathcal{L}_{\mathrm{mask}}(\mathbf{m}^{u}, \mathbf{m}^{p})~,
\end{equation}
where $\mathcal{L}_{\mathrm{mask}}$ is the binary cross-entropy loss and $\lambda$ controls its relative weight. At inference, a Gaussian sample $\mathbf{\hat{x}}_{T}^{u} \sim \mathcal{N}(\mathbf{0}, \mathbf{I})$ is iteratively denoised to produce a structured range image $\mathbf{\hat{x}}_{0}^{u}$ along with occupancy mask $\mathbf{\hat{m}}^{p}$, which constrains generation within valid spatial regions and preserves the sparsity pattern of uncertainty scenes.

\subsection{Uncertainty-Conditioned Scene Completion}
\label{subsec:scene_completion}

\noindent
While sparse uncertainty scenes emphasize semantically ambiguous regions or structurally unstable regions, they only capture partial geometric information. To generate complete and coherent LiDAR frames, we design an uncertainty-conditioned diffusion model that synthesizes full scenes under the guidance of these uncertainty priors.

\noindent\textbf{Problem Formulation.}
Given a sparse uncertainty range image $\mathbf{x}_{0}^{u}$, the goal is to reconstruct a dense LiDAR observation $\mathbf{x}_{0} \in \mathbb{R}^{H \times W \times 2}$. The conditional diffusion model $\epsilon_{\theta}^{c}$ is trained to approximate the conditional distribution $p(\mathbf{x}_{0} \mid \mathbf{x}_{0}^{u})$, enabling the model to complete the scene while respecting the structural cues encoded in $\mathbf{x}_{0}^{u}$.

\noindent\textbf{Conditional Diffusion Process.}
Similar to the unconditional diffusion defined in \cref{eq:forward}, the forward process gradually perturbs the complete scene $\mathbf{x}_{0}$:
\begin{equation}
    q(\mathbf{x}_{t} \mid \mathbf{x}_{0}) = \mathcal{N}(\mathbf{x}_{t}; \sqrt{\bar{\alpha}_{t}} \mathbf{x}_{0}, (1 - \bar{\alpha}_{t}) \mathbf{I})~.
\end{equation}
During denoising, the model integrates uncertainty priors as conditional input: $\epsilon_{\theta}^{c}: (\mathbf{x}_{t}, t, \mathbf{x}_{0}^{u}) \rightarrow \epsilon^{c}$. We concatenate $\mathbf{x}_{t}$ and $\mathbf{x}_{0}^{u}$ along the feature dimension, allowing the network to leverage both global and local cues from uncertainty regions to predict the denoising direction at each timestep. This explicit conditioning encourages structurally plausible and semantically consistent reconstruction.

\noindent\textbf{Training Objective.}
The model is optimized with the standard conditional diffusion loss:
\begin{equation}
    \mathcal{L}_{c} = \mathbb{E}_{t, \mathbf{x}_{0}, \epsilon^{c}} \left[ \| \epsilon^{c} - \epsilon_{\theta}^{c}(\mathbf{x}_{t}, t, \mathbf{x}_{0}^{u}) \|_{2}^{2} \right]~.
\end{equation}
By conditioning on the uncertainty prior, $\epsilon_{\theta}^{c}$ learns to treat these regions as structural anchors, ensuring accurate completion of occluded, distant, or small-scale objects while maintaining global scene coherence.

\noindent\textbf{Inference.}
At inference, a Gaussian sample $\mathbf{\hat{x}}_{T} \sim \mathcal{N}(\mathbf{0}, \mathbf{I})$ is iteratively denoised conditioned on $\mathbf{\hat{x}}_{0}^{u}$, producing a fully reconstructed LiDAR frame $\mathbf{\hat{x}}_{0}$. The uncertainty-guided conditioning enforces spatial fidelity in challenging regions, resulting in complete scenes that are both geometrically consistent and semantically plausible.

\begin{figure}[t]
    \centering
    \includegraphics[width=\linewidth]{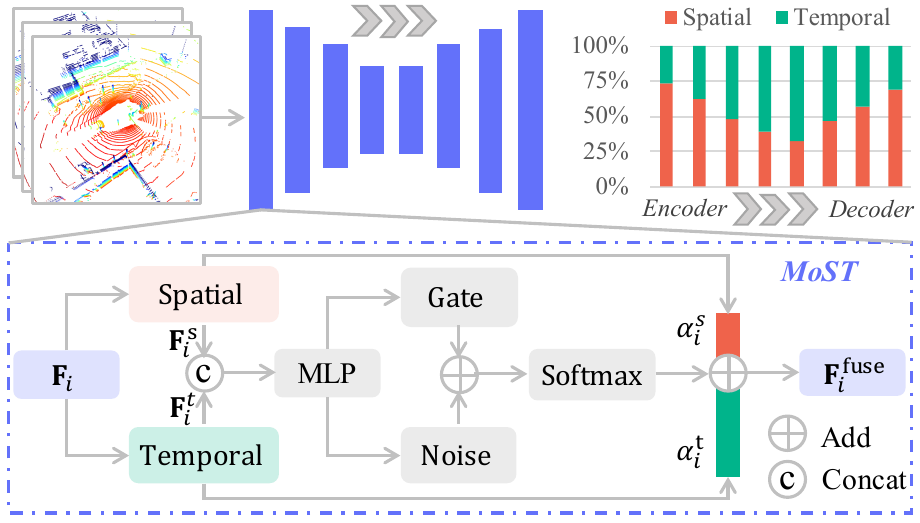}
    \vspace{-0.6cm}
    \caption{Illustration of the \textbf{Mixture of Spatio-Temporal (MoST)} block. It decomposes features along spatial and temporal dimensions and adaptively fuses them to maintain both spatial fidelity and temporal coherence. Near the network input and output, MoST emphasizes spatial cues, while in intermediate layers it focuses more on temporal dynamics.}
    \vspace{-0.4cm}
    \label{fig:most}
\end{figure}

\subsection{Mixture of Spatio-Temporal}
\label{subsec:most}

\noindent
Temporal coherence is essential for modeling dynamic real-world environments. While prior methods~\cite{nakashima2024r2dm,liu2026veila,nakashima2025r2flow} focus on spatial reconstruction, maintaining consistent temporal evolution across frames remains challenging in dynamic modeling. To jointly ensure spatial fidelity and temporal consistency, we propose the \textbf{M}ixture \textbf{o}f \textbf{S}patio-\textbf{T}emporal (\textbf{MoST}) block, as illustrated in \cref{fig:most}. MoST is a core diffusion component that simultaneously captures fine-grained spatial geometry within each frame and smooth temporal transitions between frames, producing LiDAR sequences that are both geometrically accurate and temporally stable.

\noindent\textbf{Feature Decomposing.}
Given the intermediate features $\mathbf{F}_{i} \in \mathbb{R}^{C_{i} \times L \times H_{i} \times W_{i}}$, where $L$ denotes the temporal length, $(H_{i}, W_{i})$ represent the spatial resolutions, and $C_{i}$ is the channel dimension, MoST decomposes $\mathbf{F}_{i}$ into two parallel branches: a \textit{spatial branch} that captures intra-frame geometric details through spatial convolution, producing $\mathbf{F}_{i}^{s}$, and a \textit{temporal branch} that models inter-frame consistency via temporal convolution, yielding $\mathbf{F}_{i}^{t}$, with both branches having the same output shape of $C_{i} \times L \times H_{i} \times W_{i}$.

\noindent\textbf{Spatio-Temporal Adaptive Fusion.}
To adaptively balance spatial and temporal cues, the two branch features are concatenated and refined through a lightweight Multi-Layer Perceptron (MLP) for shared embedding:
\begin{equation}
    \mathbf{F}_{i}^{\mathrm{share}} = \texttt{MLP}([\mathbf{F}_{i}^{s}; \mathbf{F}_{i}^{t}])~,
\end{equation}
where $[\cdot~;~\cdot]$ denotes feature concatenation. Inspired by mixture-of-experts designs~\cite{xu2025limoe}, we introduce a gating mechanism that dynamically adjusts the contribution of spatial and temporal activations. To enhance robustness and avoid deterministic overfitting, a stochastic noise module perturbs the gating process during training:
\begin{equation}
    (\alpha_{i}^{s}, \alpha_{i}^{t}) = \texttt{Softmax}(\mathbf{F}_{i}^{\mathrm{share}} \cdot \mathbf{W}_{i}^{g} + \mathbb{I}(\chi \cdot \sigma(\mathbf{F}_{i}^{\mathrm{share}} \cdot \mathbf{W}_{i}^{z})))~,
\end{equation}
where $\mathbf{W}_{i}^{g}, \mathbf{W}_{i}^{z} \in \mathbb{R}^{C_{i} \times 2}$ are trainable gating and noise-projection matrices, $\chi \sim \mathcal{N}(\mathbf{0}, \mathbf{I})$ denotes Gaussian noise that stochastically perturbs feature activations, $\sigma(\cdot)$ is the Softplus activation ensuring smooth and positive modulation, and $\mathbb{I}(\cdot)$ activates the noise term only during training. The normalized gating weights $(\alpha_{i}^{s}, \alpha_{i}^{t})$ are then used to adaptively fuse the spatial and temporal features:
\begin{equation}
    \mathbf{F}_{i}^{\mathrm{fuse}} = \alpha_{i}^{s} \odot \mathbf{F}_{i}^{s} + \alpha_{i}^{t} \odot \mathbf{F}_{i}^{t}~,
\end{equation}
where $\odot$ denotes element-wise multiplication. This adaptive fusion dynamically modulates the contributions from spatial and temporal representations, producing semantically rich and temporally consistent features.

\noindent\textbf{Weight Regularization.}
To prevent the gating mechanism from overemphasizing a single modality, we introduce a regularization term encouraging balanced utilization of spatial and temporal cues:
\begin{equation}
\mathcal{L}_{\mathrm{reg}, i} = \frac{\mathrm{Var}(\alpha_{i}^{s})}{(\mathbb{E}[\alpha_{i}^{s}])^{2}} + \frac{\mathrm{Var}(\alpha_{i}^{t})}{(\mathbb{E}[\alpha_{i}^{t}])^{2}}~,
\end{equation}
where $\mathrm{Var}(\cdot)$ and $\mathbb{E}[\cdot]$ denote variance and expectation, respectively. This term stabilizes the attention weights, ensuring MoST effectively leverages both spatial geometry and temporal dynamics without bias toward either dimension.

\noindent\textbf{Role in Our Framework.}
MoST serves as a fundamental building block of the diffusion network. It simultaneously captures detailed spatial geometry within each LiDAR frame while modeling smooth temporal transitions across consecutive frames. By adaptively fusing spatial and temporal features at multiple stages of the network, MoST enables the generation of LiDAR sequences that are both geometrically precise and temporally coherent, ensuring realistic and consistent scene synthesis over time.

\section{Experiments}
\label{sec:experiments}

\begin{figure*}[t]
    \centering
    \includegraphics[width=\linewidth]{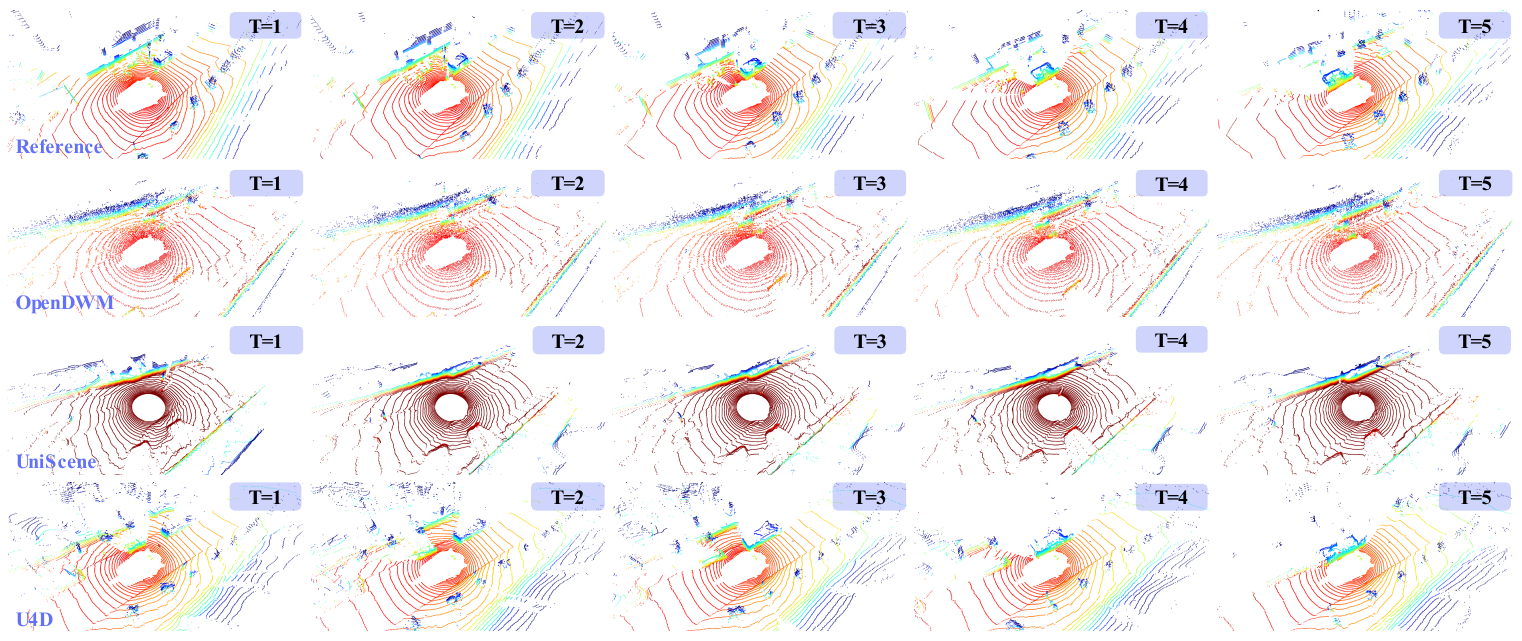}
    \vspace{-0.6cm}
    \caption{Qualitative results of \textbf{sequence point cloud generation} on the \textit{nuScenes} dataset~\cite{caesar2020nuscenes}. U4D preserves both geometric fidelity and temporal consistency, producing sequences most similar to the reference. It reliably reconstructs distant, sparse regions and captures dynamic objects across frames, maintaining coherent structure and motion. Frames are shown in temporal order from left to right. The colors are rendered based on the height information of the point cloud. Best viewed in zoom.}
    \label{fig:vis_scene}
    \vspace{-0.4cm}
\end{figure*}

\subsection{Configurations}

\noindent\textbf{Benchmarking Details.}
We conduct benchmarking experiments on the nuScenes~\cite{caesar2020nuscenes} and SemanticKITTI~\cite{behley2019semantickitti} datasets, where the range image resolutions are set to $32 \times 1024$ and $64\times 1024$, respectively. All experiments are implemented using the PyTorch framework~\cite{paszke2019pytorch} on $4$ NVIDIA RTX $4090$ GPUs with a total batch size of $8$. Following the baseline R2DM~\cite{nakashima2024r2dm}, the diffusion model is trained for $500{,}000$ steps using the AdamW optimizer~\cite{loshchilov2019adamw} with an initial learning rate of $1{\times}10^{-4}$. The learning rate follows a cosine annealing schedule with a warm-up phase during the first $10{,}000$ steps~\cite{liu2022cosine}. During inference, the diffuser performs $256$ denoising steps to ensure a fair comparison with the baseline R2DM~\cite{nakashima2024r2dm}. Please refer to the Appendix for more details.

\noindent\textbf{Evaluation Metrics.}
We assess the quality of generated LiDAR scenes from three perspectives: (1) \textit{Geometric and Spatial Fidelity.} We adopt the \textbf{Fr\'echet Range Distance (FRD)}, \textbf{Fr\'echet Point Distance (FPD)}, \textbf{Jensen-Shannon Divergence (JSD)} and \textbf{Maximum Mean Discrepancy (MMD)} on range images, point clouds, and bird's-eye-view (BEV)~\cite{nakashima2024r2dm,liu2026veila}. (2) \textit{Temporal Coherence.} Following~\cite{liang2026lidarcrafter}, we use the \textbf{Temporal Transformation Consistency Error (TTCE)} and \textbf{Chamfer Temporal Consistency (CTC)} metrics to measure frame-to-frame alignment accuracy and overall motion stability across sequences. (3) \textit{Downstream Utility.} For assessing the usefulness of generated data in perception tasks, we evaluate \textbf{Mean Intersection-over-Union (mIoU)} for semantic segmentation and \textbf{Expected Calibration Error (ECE)}~\cite{kong2025calib3d} for measuring the calibration reliability of uncertainty-aware segmentation models.

\begin{table}[t]
    \centering
    \caption{\textbf{Comparison of state-of-the-art LiDAR scene generation methods} on the \textit{nuScenes}~\cite{caesar2020nuscenes} dataset. Metrics marked with $\downarrow$ indicate that lower values are better. The \textbf{MMD} scores are reported in units of $10^{-4}$. The \textbf{best} and \underline{second-best} scores are highlighted in \textbf{bold} and \underline{underline}, respectively.}
    \vspace{-0.3cm}
    \label{tab:benchmark_nus}
    \resizebox{\linewidth}{!}{
    \begin{tabular}{r|r|cccc}
        \toprule
        \textbf{Method} & \textbf{Venue} & \textbf{FRD}~$\downarrow$ & \textbf{FPD}~$\downarrow$ & \textbf{JSD}~$\downarrow$ & \textbf{MMD}~$\downarrow$
        \\\midrule\midrule
        LiDARGen~\cite{zyrianov2022lidargen} & ECCV'22 & $549.18$ & $22.80$ & \underline{$0.04$} & $0.76$
        \\
        LiDM~\cite{ran2024lidm} & CVPR'24 & - & $30.77$ & $0.07$ & $3.86$
        \\
        R2DM~\cite{nakashima2024r2dm} & ICRA'24 & \underline{$253.80$} & \underline{$14.35$} & $\mathbf{0.03}$ & $\mathbf{0.48}$
        \\
        Text2LiDAR~\cite{wu2024text2lidar} & ECCV'24 & $953.18$ & $147.48$ & $0.09$ & $12.50$
        \\
        UniScene~\cite{li2025uniscene} & CVPR'25 & - & $976.47$ & $0.32$ & $13.61$
        \\
        OpenDWM~\cite{opendwm} & CVPR'25 & - & $714.19$ & $0.20$ & $5.61$
        \\\midrule
        \cellcolor{gen_blue!12}\textsl{U4D} & \cellcolor{gen_blue!12}\textbf{Ours} & \cellcolor{gen_blue!12}$\mathbf{223.96}$ & \cellcolor{gen_blue!12}$\mathbf{12.90}$ & \cellcolor{gen_blue!12}$\mathbf{0.03}$ & \cellcolor{gen_blue!12}\underline{$0.53$}
        \\\bottomrule
    \end{tabular}}
    \vspace{-0.4cm}
\end{table}

\subsection{Comparative Study}

\noindent\textbf{Scene-Level Fidelity.}
We first benchmark U4D against state-of-the-art LiDAR scene generation methods from a scene-level spatial fidelity perspective. Specifically, we sample $10{,}000$ sequences and evaluate the first frame of each sequence to ensure a fair comparison with single-frame generation baselines producing the same number of scenes. As shown in \cref{tab:benchmark_nus} and \cref{tab:benchmark_semkitti}, U4D consistently outperforms existing methods, achieving an FRD of $223.96$ and an FPD of $12.90$ on nuScenes~\cite{caesar2020nuscenes}, and $245.73$ (FRD) and $10.92$ (FPD) on SemanticKITTI~\cite{behley2019semantickitti}, surpassing R2DM~\cite{nakashima2024r2dm} by approximately $6\%$-$11\%$. For BEV-based metrics, including JSD and MMD, U4D also achieves competitive or superior performance, demonstrating robust spatial consistency across viewpoints. These results highlight U4D’s strong ability to generate geometrically accurate and perceptually consistent LiDAR scenes.

\begin{table}[t]
    \centering
    \caption{\textbf{Comparison of state-of-the-art LiDAR scene generation methods} on the \textit{SemanticKITTI}~\cite{behley2019semantickitti} dataset. Metrics marked with $\downarrow$ indicate that lower values are better. The \textbf{MMD} scores are reported in units of $10^{-4}$. The \textbf{best} and \underline{second-best} scores are highlighted in \textbf{bold} and \underline{underline}, respectively.}
    \vspace{-0.3cm}
    \label{tab:benchmark_semkitti}
    \resizebox{\linewidth}{!}{
    \begin{tabular}{r|r|cccc}
        \toprule
        \textbf{Method} & \textbf{Venue} & \textbf{FRD}~$\downarrow$ & \textbf{FPD}~$\downarrow$ & \textbf{JSD}~$\downarrow$ & \textbf{MMD}~$\downarrow$
        \\\midrule\midrule
        LiDARGen~\cite{zyrianov2022lidargen} & ECCV'22 & $735.49$ & $119.69$ & $0.13$ & $21.90$
        \\
        LiDM~\cite{ran2024lidm} & CVPR'24 & - & $496.78$ & $0.08$ & $9.20$
        \\
        R2DM~\cite{nakashima2024r2dm} & ICRA'24 & \underline{$262.85$} & \underline{$12.06$} & $\mathbf{0.03}$ & \underline{$0.89$}
        \\
        Text2LiDAR~\cite{wu2024text2lidar} & ECCV'24 & $567.47$ & $16.78$ & $0.08$ & $4.24$
        \\\midrule
        \cellcolor{gen_blue!12}\textsl{U4D} & \cellcolor{gen_blue!12}\textbf{Ours} & \cellcolor{gen_blue!12}$\mathbf{245.73}$ & \cellcolor{gen_blue!12}$\mathbf{10.92}$ & \cellcolor{gen_blue!12}\underline{$0.04$} & \cellcolor{gen_blue!12}$\mathbf{0.85}$
        \\\bottomrule
    \end{tabular}}
    \vspace{-0.25cm}
\end{table}

\begin{table}[t]
    \centering
    \caption{\textbf{Comparison of temporal consistency} in 4D LiDAR scene generation on the \textit{nuScenes}~\cite{caesar2020nuscenes} dataset. Metrics marked with $\downarrow$ indicate that lower values are better. The \textbf{best} and \underline{second-best} scores are highlighted in \textbf{bold} and \underline{underline}, respectively. Numbers denote frame intervals.}
    \vspace{-0.3cm}
    \label{tab:benchmark_temporal}
    \resizebox{\linewidth}{!}{
    \begin{tabular}{r|r|cc|cccc}
        \toprule
        \multirow{2}{*}{\textbf{Method}} & \multirow{2}{*}{\textbf{Venue}} & \multicolumn{2}{c|}{\textbf{TTCE}~$\downarrow$} & \multicolumn{4}{c}{\textbf{CTC}~$\downarrow$}
        \\
        & & $\mathbf{3}$ & $\mathbf{4}$ & $\mathbf{1}$ & $\mathbf{2}$ & $\mathbf{3}$ & $\mathbf{4}$
        \\\midrule\midrule
        UniScene~\cite{li2025uniscene} & CVPR'25 & $2.74$ & $3.69$ & $\mathbf{0.90}$ & $\mathbf{1.84}$ & $3.64$ & $\mathbf{3.90}$
        \\
        OpenDWM~\cite{opendwm} & CVPR'25 & $2.68$ & $3.65$ & $1.02$ & $2.02$ & $3.37$ & $5.05$
        \\
        LiDARCrafter~\cite{liang2026lidarcrafter} & AAAI'26 & \underline{$2.65$} & \underline{$3.56$} & $1.12$ & $2.38$ & \underline{$3.02$} & $4.81$
        \\\midrule
        \cellcolor{gen_blue!12}\textsl{U4D} & \cellcolor{gen_blue!12}\textbf{Ours} & \cellcolor{gen_blue!12}$\mathbf{2.63}$ & \cellcolor{gen_blue!12}$\mathbf{3.51}$ & \cellcolor{gen_blue!12}\underline{$0.97$} & \cellcolor{gen_blue!12}\underline{$1.93$} & \cellcolor{gen_blue!12}$\mathbf{2.98}$ & \cellcolor{gen_blue!12}\underline{$4.41$}
        \\\bottomrule
    \end{tabular}}
    \vspace{-0.4cm}
\end{table}

\noindent\textbf{Temporal Coherence.}
Maintaining temporal coherence is crucial for sequential LiDAR generation, as inconsistent frame-to-frame predictions can lead to unrealistic scene dynamics. In \cref{tab:benchmark_temporal}, we evaluate U4D against recent methods, including UniScene~\cite{li2025uniscene}, OpenDWM~\cite{opendwm}, and LiDARCrafter~\cite{liang2026lidarcrafter}, on sequences sampled at $2$Hz. TTCE measures deviations between predicted and ground-truth transformations via point cloud registration, while CTC computes Chamfer distances between consecutive frames. U4D consistently achieves the lowest TTCE scores across all frame intervals and maintains competitive CTC scores, reflecting its ability to generate temporally stable sequences with smooth and realistic motion patterns. These results highlight the effectiveness of the MoST block in capturing both spatial and temporal dependencies within sequences.

\begin{figure*}[t]
    \centering
    \includegraphics[width=\linewidth]{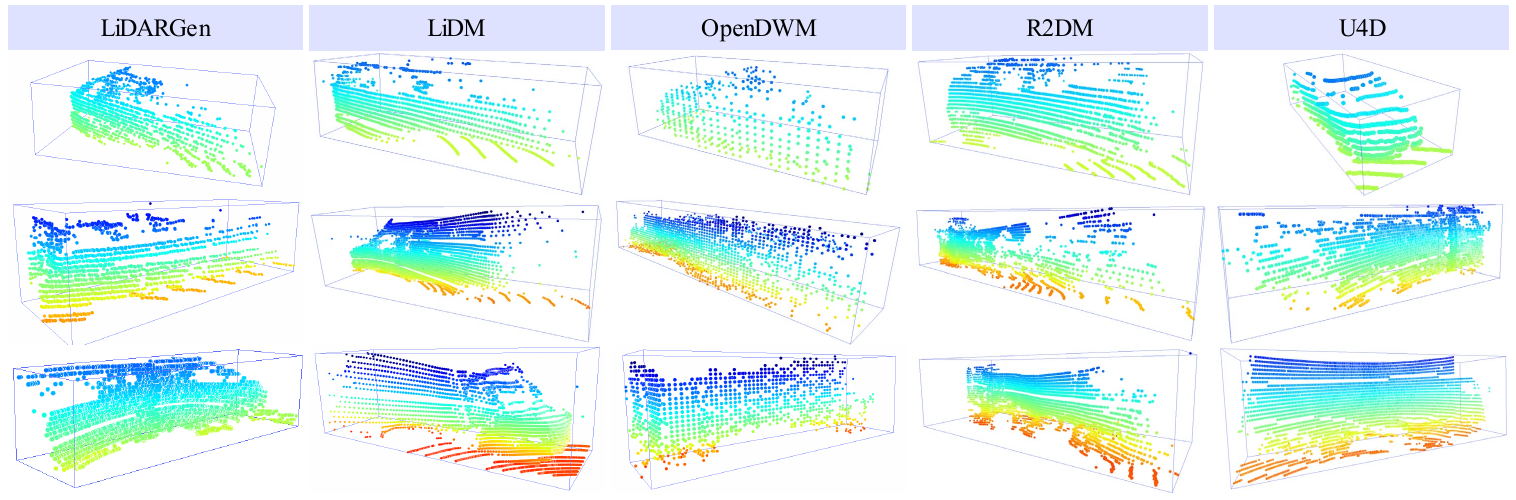}
    \vspace{-0.6cm}
    \caption{Qualitative results of \textbf{generated objects within scenes} on the \textit{nuScenes} dataset~\cite{caesar2020nuscenes}. U4D accurately captures object geometry and preserves fine structural details while maintaining realistic spatial relationships within the scene. From top to bottom are ``car'', ``bus'', and ``truck'', respectively. All objects are detected using a pretrained PointPillars~\cite{lang2019pointpillars} detector.}
    \label{fig:vis_object}
    \vspace{-0.4cm}
\end{figure*}

\begin{table}[!t]
    \centering
    \caption{\textbf{Comparison of state-of-the-art methods} on the downstream task of LiDAR semantic segmentation on the \textit{val} set of the \textit{nuScenes}~\cite{caesar2020nuscenes} dataset. The voxel- and fusion-based representations are built upon MinkUNet~\cite{choy2019minkunet} and SPVCNN~\cite{tang2020spvcnn} as backbones, respectively. The mIoU scores are reported in percentage (\%). The \textbf{best} and \underline{second-best} results within each data split and representation are highlighted in \textbf{bold} and \underline{underline}, respectively.}
    \vspace{-0.3cm}
    \label{tab:downstream_segmentation}
    \resizebox{\linewidth}{!}{
    \begin{tabular}{c|r|r|cccc}
        \toprule
        \multirow{2}{*}{\textbf{Repr.}} & \multirow{2}{*}{\textbf{Method}} & \multirow{2}{*}{\textbf{Venue}} & \multicolumn{4}{c}{\textbf{nuScenes}}
        \\
        & & & $\mathbf{1\%}$ & $\mathbf{10\%}$ & $\mathbf{20\%}$ & $\mathbf{50\%}$
        \\\midrule\midrule
        \multirow{6}{*}{\rotatebox{90}{\textbf{Voxel}}} & \cellcolor{gen_gray!20}\textit{Sup.-only} & \cellcolor{gen_gray!20}- & \cellcolor{gen_gray!20}$58.3$ & \cellcolor{gen_gray!20}$71.0$ & \cellcolor{gen_gray!20}$73.0$ & \cellcolor{gen_gray!20}$75.1$
        \\\cmidrule{2-7}
        & MeanTeacher~\cite{tarvainen2017meanteacher} & NeurIPS'17 & $60.1$ & $71.7$ & $73.4$ & $75.2$
        \\
        & LaserMix~\cite{kong2023lasermix} & CVPR'23 & $62.8$ & \underline{$73.6$} & \underline{$74.8$} & \underline{$76.1$}
        \\
        & R2DM~\cite{nakashima2024r2dm} & ICRA'24 & \underline{$64.1$} & $73.0$ & $74.3$ & $75.9$
        \\\cmidrule{2-7}
        & \cellcolor{gen_blue!12}\textsl{U4D} & \cellcolor{gen_blue!12}\textbf{Ours} & \cellcolor{gen_blue!12}$\mathbf{65.3}$ & \cellcolor{gen_blue!12}$\mathbf{73.7}$ & \cellcolor{gen_blue!12}$\mathbf{75.0}$ & \cellcolor{gen_blue!12}$\mathbf{76.4}$
        \\\midrule
        \multirow{6}{*}{\rotatebox{90}{\textbf{Fusion}}} & \cellcolor{gen_gray!20}\textit{Sup.-only} & \cellcolor{gen_gray!20}- & \cellcolor{gen_gray!20}$57.9$ & \cellcolor{gen_gray!20}$71.7$ & \cellcolor{gen_gray!20}$73.0$ & \cellcolor{gen_gray!20}$74.6$
        \\\cmidrule{2-7}
        & MeanTeacher~\cite{tarvainen2017meanteacher} & NeurIPS'17 & $59.4$ & $72.5$ & $73.1$ & $74.7$
        \\
        & LaserMix~\cite{kong2023lasermix} & CVPR'23 & $63.2$ & $\mathbf{74.1}$ & \underline{$74.6$} & \underline{$75.8$}
        \\
        & R2DM~\cite{nakashima2024r2dm} & ICRA'24 & \underline{$64.6$} & \underline{$72.7$} & $74.2$ & $75.4$
        \\\cmidrule{2-7}
        & \cellcolor{gen_blue!12}\textsl{U4D} & \cellcolor{gen_blue!12}\textbf{Ours} & \cellcolor{gen_blue!12}$\mathbf{65.3}$ & \cellcolor{gen_blue!12}$\mathbf{74.1}$ & \cellcolor{gen_blue!12}$\mathbf{74.9}$ & \cellcolor{gen_blue!12}$\mathbf{76.2}$
        \\\bottomrule
    \end{tabular}}
    \vspace{-0.4cm}
\end{table}

\noindent\textbf{LiDAR Semantic Segmentation.}
LiDAR semantic segmentation is a fundamental perception task in autonomous driving, yet it remains highly challenging in uncertain and sparse regions. To validate the fidelity and utility of our uncertainty-aware LiDAR scene generation, we apply the generated data to downstream segmentation tasks. The overall training pipeline follows LaserMix~\cite{kong2023lasermix}, where only a subset of real data is used as labeled samples, while the remaining real data and our generated uncertainty-aware scenes are treated as unlabeled data. As shown in \cref{tab:downstream_segmentation}, we evaluate U4D under two popular LiDAR representations -- \textit{sparse voxels}~\cite{choy2019minkunet} and \textit{multi-view fusion}~\cite{tang2020spvcnn}. The results show that incorporating scenes generated by U4D consistently enhances downstream segmentation performance across all label partitions, outperforming both semi-supervised learning methods~\cite{tarvainen2017meanteacher,kong2023lasermix} and scene generation augmentation baselines~\cite{nakashima2024r2dm}. This confirms that U4D not only generates geometrically and semantically coherent LiDAR scenes but also provides strong generalization benefits for real-world perception models.

\begin{table}[t]
    \centering
    \caption{\textbf{Expected calibration error (ECE, the lower the better) of various LiDAR semantic segmentation methods} on the \textit{validation} set of the \textit{nuScenes}~\cite{behley2019semantickitti} and \textit{SemanticKITTI}~\cite{behley2019semantickitti} datasets. The ECE scores are reported in percentage (\%).}
    \vspace{-0.3cm}
    \label{tab:uncertainty}
    \resizebox{\linewidth}{!}{
    \begin{tabular}{r|r|cc|cc}
        \toprule
        \multirow{2}{*}{\textbf{Method}} & \multirow{2}{*}{\textbf{Repr.}} & \multicolumn{2}{c|}{\textbf{nuScenes}} & \multicolumn{2}{c}{\textbf{SemKITTI}}
        \\
        & & Uncal & \cellcolor{gen_blue!12}\textsl{U4D} & Uncal & \cellcolor{gen_blue!12}\textsl{U4D}
        \\\midrule\midrule
        RangeNet++~\cite{milioto2019rangenet++} & Range & $4.57$ & \cellcolor{gen_blue!12}$2.72$ & $4.01$ & \cellcolor{gen_blue!12}$3.31$
        \\
        MinkUNet~\cite{choy2019minkunet} & Voxel & $2.50$ & \cellcolor{gen_blue!12}$2.45$ & $4.11$ & \cellcolor{gen_blue!12}$3.56$
        \\
        SPVCNN~\cite{tang2020spvcnn} & Fusion & $2.61$ & \cellcolor{gen_blue!12}$2.18$ & $3.61$ & \cellcolor{gen_blue!12}$3.05$
        \\\bottomrule
    \end{tabular}}
    \vspace{-0.6cm}
\end{table}

\noindent\textbf{Uncertainty Estimation.}
To quantitatively assess the benefit of the proposed uncertainty-aware generation framework on real-world perception, we evaluate the calibration performance of LiDAR semantic segmentation methods using the ECE metric following Calib3D~\cite{kong2025calib3d}. Specifically, we first generate pseudo-labels for the synthesized scenes using a pretrained segmentation model~\cite{xu2025frnet}, and then jointly train the model with both real and generated data. As shown in \cref{tab:uncertainty}, ``Uncal'' denotes models trained solely on real data. U4D consistently achieves lower ECE values, demonstrating that uncertainty-aware scene generation improves model calibration, reliability, and prediction confidence.

\noindent\textbf{Qualitative Results.}
\cref{fig:vis_scene} presents qualitative comparisons of sequential LiDAR scenes generated by U4D and recent sequence generation methods~\cite{li2025uniscene,opendwm}. Spatially, U4D produces more realistic and structurally faithful scenes that closely match the reference. Temporally, it maintains smooth object motion and ego-motion consistency across frames. Furthermore, \cref{fig:vis_object} visualizes generated object instances within scenes, where all objects are detected using a pretrained PointPillars~\cite{lang2019pointpillars} detector. As shown, U4D can generate objects with clear geometric boundaries, consistent scales, and coherent placement relative to surrounding structures. These advantages can be attributed to two factors. First, the proposed uncertainty-aware generation mechanism guides the model to focus more on ambiguous and structurally complex regions, thereby improving local geometry and object fidelity. Second, the MoST block adaptively fuses spatial and temporal cues, ensuring that objects remain temporally consistent and physically plausible.

\subsection{Ablation Study}

\noindent\textbf{Selection of Uncertainty Regions.}
Identifying suitable uncertainty regions is essential, as they provide crucial structural priors that guide the generation. In this ablation study, we evaluate different strategies for selecting uncertainty regions, with the results summarized in \cref{tab:abl_uncertainty}. When trained without any uncertainty conditioning (row~1), the model achieves $235.91$ FRD, $14.03$ FPD, and an ECE of $3.98$. Random sampling (row~2) provides negligible improvement in generation quality and even degrades calibration performance. In contrast, both confidence-based (row~3) and entropy-based (row~4) selection strategies lead to significant gains in scene fidelity and calibration, as they effectively localize regions of high semantic ambiguity. Among them, the entropy-based strategy achieves the best overall performance, demonstrating its superior ability to capture diverse and informative uncertainty patterns for guiding generation.

\begin{table}[t]
    \centering
    \caption{Ablation study on the \textbf{selection of uncertainty regions} on \textit{nuScenes}~\cite{caesar2020nuscenes}. Metrics marked with $\downarrow$ indicate that lower values are better. The \textbf{MMD} scores are reported in units of $10^{-4}$. ECE is evaluated using RangeNet++~\cite{milioto2019rangenet++} and reported in percentage (\%).}
    \vspace{-0.3cm}
    \label{tab:abl_uncertainty}
    \resizebox{\linewidth}{!}{
    \begin{tabular}{c|r|cccc|c}
        \toprule
        \textbf{\#} & \textbf{Metric} & \textbf{FRD}~$\downarrow$ & \textbf{FPD}~$\downarrow$ & \textbf{JSD}~$\downarrow$ & \textbf{MMD}~$\downarrow$ & \textbf{ECE}~$\downarrow$
        \\\midrule\midrule
        (1) & None & $235.91$ & $14.03$ & $0.04$ & $0.65$ & $3.98$
        \\
        (2) & Random & $235.23$ & $13.21$ & $0.04$ & $0.55$ & $4.35$
        \\
        (3) & Confidence & $228.24$ & $13.04$ & $\mathbf{0.03}$ & $0.72$ & $3.02$
        \\\midrule
        \cellcolor{gen_blue!12}(4) & \cellcolor{gen_blue!12}Entropy & \cellcolor{gen_blue!12}$\mathbf{223.96}$ & \cellcolor{gen_blue!12}$\mathbf{12.90}$ & \cellcolor{gen_blue!12}$\mathbf{0.03}$ & \cellcolor{gen_blue!12}$\mathbf{0.53}$ &
        \cellcolor{gen_blue!12}$\mathbf{2.72}$
        \\\bottomrule
    \end{tabular}}
    \vspace{-0.25cm}
\end{table}

\begin{table}[t]
    \centering
    \caption{Ablation study on the \textbf{design of the MoST block} on the \textit{nuScenes}~\cite{caesar2020nuscenes} dataset. Metrics marked with $\downarrow$ indicate that lower values are better. The \textbf{MMD} scores are reported in units of $10^{-4}$.}
    \vspace{-0.3cm}
    \label{tab:abl_most}
    \resizebox{\linewidth}{!}{
    \begin{tabular}{c|cc|cccc}
        \toprule
        \textbf{\#} & \textbf{Parallel} & \textbf{Fusion} & \textbf{FRD}~$\downarrow$ & \textbf{FPD}~$\downarrow$ & \textbf{JSD}~$\downarrow$ & \textbf{MMD}~$\downarrow$
        \\\midrule\midrule
        (1) & \textcolor{gen_red}{\xmark} & None & $536.23$ & $23.34$ & $0.63$ & $2.43$
        \\
        (2) & \textcolor{gen_blue}{\cmark} & Addition & $242.81$ & $13.42$ & $0.28$ & $0.65$
        \\
        (3) & \textcolor{gen_blue}{\cmark} & Concat & $242.43$ & $\mathbf{12.51}$ & $\mathbf{0.03}$ & $0.68$
        \\\midrule
        \cellcolor{gen_blue!12}(4) & \cellcolor{gen_blue!12}\textcolor{gen_blue}{\cmark} & \cellcolor{gen_blue!12}Adaptively & \cellcolor{gen_blue!12}$\mathbf{223.96}$ & \cellcolor{gen_blue!12}$12.90$ & \cellcolor{gen_blue!12}$\mathbf{0.03}$ & \cellcolor{gen_blue!12}$\mathbf{0.53}$
        \\\bottomrule
    \end{tabular}}
    \vspace{-0.4cm}
\end{table}

\noindent\textbf{Design of MoST Block.}
The Mixture of Spatio-Temporal (MoST) block serves as a key component of our diffusion backbone, designed to adaptively fuse spatial and temporal cues for coherent LiDAR scene generation. In this ablation, we investigate how different fusion strategies affect generation quality, as shown in \cref{tab:abl_most}. We first follow prior video generation approaches~\cite{gao2024vista} and apply spatial and temporal operations in a cascaded manner (row~1). This configuration yields suboptimal results, likely because the deeper cascaded structure hampers convergence and optimization. We then decompose features into spatial and temporal branches and fuse them in parallel using either element-wise addition (row~2) or concatenation (row~3). Both strategies significantly improve generation quality, as they expand network width rather than depth, facilitating more stable optimization. Finally, we introduce an adaptive fusion mechanism inspired by the mixture-of-experts paradigm~\cite{xu2025limoe} (row~4), where the model learns to dynamically balance spatial and temporal information. This design achieves the best generation quality, demonstrating the effectiveness of adaptive spatio-temporal fusion for coherent 4D LiDAR generation.

\noindent\textbf{Spatio-Temporal Activations.}
To further examine how the MoST block fuses spatial and temporal information across different network stages, we analyze the relative activation weights of its two branches. As shown in the top-right of \cref{fig:most}, we visualize the averaged weighting distribution of spatial and temporal branches throughout the diffusion network. We observe that near the input and output layers, the spatial branch contributes more prominently, as these stages mainly focus on reconstructing local geometric details and structural integrity of LiDAR frames. In contrast, the temporal branch exhibits stronger activations in intermediate layers, where the model captures motion dynamics and ensures temporal consistency across frames. This observation validates our design intuition that spatial cues dominate at the boundaries for geometric fidelity, while temporal cues become essential in the latent space to model scene evolution and motion continuity. The adaptive allocation of activations allows MoST to balance geometric reconstruction and motion modeling, enabling the diffusion network to generate LiDAR sequences that are both spatially accurate and temporally coherent.

\begin{table}[t]
    \centering
    \caption{Ablation study on the \textbf{efficiency of generative models}. The table reports the average inference time per frame (\textbf{I.T.}, in seconds). ``DM'' refers to the diffusion model.}
    \vspace{-0.3cm}
    \label{tab:abl_efficacy}
    \resizebox{\linewidth}{!}{
    \begin{tabular}{r|r|c|c|c}
        \toprule
        \textbf{Method} & \textbf{Venue} & \textbf{Generator} & \textbf{Architecture} & \textbf{I.T.}
        \\\midrule\midrule
        LiDARGen~\cite{zyrianov2022lidargen} & ECCV'22 & Energy-Based & U-Net & $67$
        \\
        LiDM~\cite{ran2024lidm} & CVPR'24 & Latent DM & Curve U-Net & $3.5$
        \\
        R2DM~\cite{nakashima2024r2dm} & ICRA'24 & DM & Curve U-Net & $3.5$
        \\
        UniScene~\cite{li2025uniscene} & CVPR'25 & Volume Rendering & Sparse U-Net & $2.1$
        \\
        OpenDWM~\cite{opendwm} & CVPR'25 & Latent R & U-Net & $12.1$
        \\\midrule
        \cellcolor{gen_blue!12}\textbf{U4D} & \cellcolor{gen_blue!12}\textbf{Ours} & \cellcolor{gen_blue!12}\textbf{DM} & \cellcolor{gen_blue!12}\textbf{Curve U-Net} & \cellcolor{gen_blue!12}$8.9$
        \\\bottomrule
    \end{tabular}}
    \vspace{-0.4cm}
\end{table}

\noindent\textbf{Efficacy Analysis.}
\cref{tab:abl_efficacy} summarizes the efficiency of U4D compared with other generative models. Compared with single-frame generators~\cite{ran2024lidm,nakashima2024r2dm}, U4D requires slightly longer inference time since it generates multiple frames simultaneously, which inevitably increases computational demand. However, when compared to other sequential generative models, U4D demonstrates competitive efficiency while achieving superior generation quality. Its inference speed is slower than UniScene~\cite{li2025uniscene}, primarily due to the higher spatial resolution of our range-based representation. These observations highlight the potential of U4D as a strong yet scalable baseline, motivating future work toward more efficient architectures such as latent diffusion models.

\section{Conclusion}
\label{sec:conclusion}

In this work, we propose \textbf{U4D}, the first uncertainty-aware generative framework for 4D LiDAR world modeling. U4D first estimates spatial uncertainty maps from a pretrained segmentation model and introduces a two-stage diffusion pipeline that generates high-fidelity LiDAR scenes in a ``hard-to-easy'' manner: (1) the \textit{uncertainty-region modeling} stage reconstructs high-entropy regions with fine geometric fidelity, and (2) the \textit{uncertainty-conditional completion} stage synthesizes the remaining areas guided by the learned structural uncertainty prior. To further ensure temporal coherence, U4D incorporates a Mixture of Spatio-Temporal (MoST) block that adaptively modulates spatial and temporal cues, enabling geometrically precise and temporally consistent LiDAR sequences. Extensive experiments validate the effectiveness of U4D in uncertainty estimation and real-world downstream applications, paving the way toward more reliable 4D LiDAR world models.

\section*{Acknowledgments}

This work was supported in part by the Natural Science Foundation of China under Grant U24B20155.

This research is supported by cash and in-kind funding from NTU S-Lab and industry partner(s). This study is also supported by the Ministry of Education, Singapore, under its MOE AcRF Tier 2 (MOE-T2EP20221-0012, MOE-T2EP20223-0002). Lingdong Kong is supported by the Apple Scholars in AI/ML Ph.D. Fellowship program.

\section*{Appendix}
\startcontents[appendices]
\printcontents[appendices]{l}{1}{\setcounter{tocdepth}{3}}
\vspace{0.2cm}

\section{Additional Implementation Details}
\label{sec_supp:implementation}

In this section, we provide comprehensive implementation details to facilitate reproducibility and enable understanding of our experimental setup. We elaborate on the training pipeline, network configurations, data preprocessing, and evaluation protocols adopted throughout our experiments.

\subsection{Training Configurations}
\label{subsec_supp:training}

In this section, we provide detailed training configurations to facilitate the reproducibility of U4D.

\vspace{0.1cm}
\noindent\textbf{Data Preprocessing.}
As described in the main paper, we first convert raw LiDAR point clouds into range-image representations~\cite{kong2023rangeformer,milioto2019rangenet++,kong2023robo3d}, which serve as the input format for our diffusion-based generative framework. Given a 3D point $(x_{i}, y_{i}, z_{i})$, its projected pixel coordinate $(u_{i}, v_{i})$ in a range image of resolution $H \times W$ is computed as:
\begin{equation}
    \label{eq:range_projection}
    \binom{u_{i}}{v_{i}} = \binom{\frac{1}{2}\left[ 1 - \arctan(y_{i}, x_{i})\pi^{-1} \right] W}{\left[ 1 - (\arcsin(z_{i} d_{i}^{-1}) - f_{\mathrm{down}}) f^{-1} \right] H}~,
\end{equation}
where $d_{i} = \sqrt{x_{i}^{2} + y_{i}^{2} + z_{i}^{2}}$ denotes the radial distance of the point, and $f = f_{\mathrm{up}} - f_{\mathrm{down}}$ is the vertical field of view (FOV) spanned by the LiDAR sensor. The parameters $f_{\mathrm{up}}$ and $f_{\mathrm{down}}$ correspond to the upper and lower elevation angles defined by the sensor. For range-image construction, we adopt resolutions $(H, W) = (32, 1024)$ for nuScenes~\cite{caesar2020nuscenes} and $(64, 1024)$ for SemanticKITTI~\cite{behley2019semantickitti}. The vertical FOV settings $(f_{\mathrm{down}}, f_{\mathrm{up}})$ follow the respective sensor configurations of each dataset: $(-30^{\circ}, 10^{\circ})$ for nuScenes~\cite{caesar2020nuscenes} and $(-25^{\circ}, 3^{\circ})$ for SemanticKITTI~\cite{behley2019semantickitti}.

Each resulting range image $\mathbf{X} \in \mathbb{R}^{H \times W \times 2}$ contains two channels -- depth and intensity\footnote{For the nuScenes~\cite{caesar2020nuscenes} dataset, the raw intensity values lies in $[0, 255]$ and are normalized to $[0, 1]$ by dividing by $255.0$.}. To stabilize diffusion training, we apply depth compression and channel-level normalization before feeding data to the model. For the depth channel, we employ a logarithmic compression~\cite{nakashima2024r2dm}:
\begin{equation}
    \label{eq:logarithmic}
    \mathbf{X}_{d}^{\mathrm{norm}} = \frac{\log_{2}(\mathbf{X}_{d} + 1)}{\log_{2}(d_{\mathrm{max}} + 1)}~,
\end{equation}
where $d_{\mathrm{max}}$ is the maximum measurable LiDAR range, set to $80.0$ meters in all experiments. This compression mitigates the large dynamic range of raw depth value and enhances the stability of diffusion noise prediction, particularly in distant sparse regions. Then, all pixel values are linearly scaled into the range $[-1, 1]$ as:
\begin{equation}
    \label{eq:scale}
    \mathbf{X}^{\mathrm{input}} = 2 \cdot \mathbf{X}^{\mathrm{norm}} - 1~.
\end{equation}
This standardization step ensures consistent input statistics across datasets and improves training convergence.

\vspace{0.1cm}
\noindent\textbf{Network Architectures.}
Following R2DM~\cite{nakashima2024r2dm}, we adopt a $4$-layer Efficient U-Net~\cite{saharia2022efficient-unet} as the backbone of our diffusion network, with intermediate feature dimensions of $64$, $128$, $256$, and $512$. Each layer contains three Mixture of Spatio-Temporal (MoST) blocks, and the spatial resolution is downsampled by a factor of two along both the vertical and horizontal dimensions in the last three layers. In a MoST block, the input features are denoted as $\mathbf{F}_{i} \in \mathbb{R}^{C_{i} \times L \times H_{i} \times W_{i}}$, where $C_{i}$ is the channel dimension, $L$ is the temporal length, and $(H_{i}, W_{i})$ are the spatial resolutions at the $i$-th layer. The spatial branch processes $\mathbf{F}_{i}$ using a $1 \times 3 \times 3$ convolution to encode intra-frame geometric structures, while the temporal branch applies a $3 \times 1 \times 1$ convolution to capture inter-frame temporal dependencies. The outputs of the two branches are fused and added back to the input through a residual connection, ensuring stable training and efficient integration of spatial and temporal cues throughout the diffusion process.

\vspace{0.1cm}
\noindent\textbf{Selection of Uncertainty Regions.}
To determine the uncertainty regions, we employ a pretrained RangeNet++~\cite{milioto2019rangenet++} semantic segmentation model to estimate the per-pixel class probability distribution. For each point, we compute its Shannon Entropy~\cite{shannon1948entropy} as:
\begin{equation}
    H(\mathbf{p}) = \sum_{c=1}^{C} D(c \mid \mathbf{p}) \log D(c \mid \mathbf{p})~,
\end{equation}
where $D(c \mid \mathbf{p})$ denotes the predicted probability of class $c$ for point $\mathbf{p}$, and $C$ is the total number of semantic classes. Points with higher entropy represent regions with greater semantic ambiguity, typically corresponding to object boundaries, distant structures, or sparsely scanned areas. To ensure sparse yet consistent uncertainty-mask coverage across the dataset, we retain the top-$20\%$ highest-entropy points for nuScenes~\cite{caesar2020nuscenes} and the top-$5\%$ for SemanticKITTI~\cite{behley2019semantickitti}. These ratios are chosen to provide sufficient supervisory signal for ambiguous regions while preserving stable coverage across frames and sequences.

\vspace{0.1cm}
\noindent\textbf{Training Hyperparameters.}
We employ a two-stage training pipeline for U4D using the PyTorch~\cite{paszke2019pytorch} framework. The first stage, \textit{uncertainty-region modeling}, is trained for $1{,}000{,}000$ steps, whereas the second stage, \textit{uncertainty-conditioned completion}, is trained for $500{,}000$ steps. Both stages use a batch size of $8$ with a sequence length of $6$. We use the AdamW optimizer~\cite{loshchilov2019adamw} with a learning rate of $1\times10^{-4}$, $\beta_1=0.9$, $\beta_2=0.99$, and $\epsilon=1\times10^{-8}$. The learning rate follows a cosine annealing schedule with a warm-up period of the first $10{,}000$ steps~\cite{liu2022cosine}. To further stabilize optimization, we apply an exponential moving average (EMA) with a decay rate of $0.995$, updated every $10$ training steps. The diffusion process adopts continuous timesteps with a cosine noise schedule, consistent with R2DM~\cite{nakashima2024r2dm}. During inference, we use $256$ sampling steps to ensure a fair comparison with R2DM~\cite{nakashima2024r2dm}. All experiments are conducted on a server equipped with four NVIDIA RTX $4090$ GPUs under mixed-precision (FP16) training.

\subsection{Evaluation Configurations}
\label{subsec_supp:evaluation}

This section summarizes the evaluation configurations and metrics used to assess the quality of LiDAR scene generation from three perspectives: $^1$\textit{Geometric and Spatial Fidelity}, $^2$\textit{Temporal Coherence}, and $^3$\textit{Downstream Utility}.

\vspace{0.1cm}
\noindent\textbf{Geometric and Spatial Fidelity.}
This set of metrics evaluates how accurately the generated LiDAR scenes capture the geometry and spatial structure of real-world environments. We adopt four measures:

\vspace{0.1cm}
\noindent$\bullet$~\textit{Fr\'echet Range Distance (FRD).}
FRD quantitatively evaluates the generation quality in the range image domain, which provides a structured 2D representation of LiDAR point clouds. Given the real set $\mathcal{R}$ and the generated set $\mathcal{G}$, their corresponding range images are processed using a RangeNet++~\cite{milioto2019rangenet++} model pretrained for semantic segmentation on the real dataset. The intermediate feature activations extracted from the backbone are denoted as $\mathcal{F}_{r}$ and $\mathcal{F}_{g}$, representing the feature distributions of real and generated scenes, respectively. The FRD is computed as:
\begin{equation}
    \label{eq:frd}
    \texttt{FRD}(\mathcal{R}, \mathcal{G}) = \| \mu_{g} - \mu_{r} \|_{2}^{2} + \texttt{Tr}\left(\Sigma_{g} + \Sigma_{r} - 2(\Sigma_{g}\Sigma_{r})^{\frac{1}{2}}\right)~,
\end{equation}
where $\mu_{r}$ and $\mu_{g}$ denote the mean feature embeddings of $\mathcal{F}_{r}$ and $\mathcal{F}_{g}$, $\Sigma_{r}$ and $\Sigma_{g}$ represent their corresponding covariance matrices, and $\texttt{{Tr}}(\cdot)$ denotes the matrix trace. A lower FRD value indicates that the generated samples more closely match real LiDAR scenes in the learned feature space, implying higher fidelity and semantic consistency.

\vspace{0.1cm}
\noindent$\bullet$~\textit{Fr\'echet Point Distance (FPD).}
While FRD operates in the range image domain, FPD assesses the generation quality directly in the 3D point cloud space, offering a complementary geometric perspective. Following~\cite{nakashima2024r2dm,liang2026lidarcrafter,liu2026lalalidar}, we employ a PointNet~\cite{qi2017pointnet} model pretrained on the ShapeNet dataset~\cite{chang2015shapenet} for $16$-class object classification to extract high-level geometric features from both the real and generated point clouds, resulting in feature sets $\mathcal{F}_{r}^{p}$ and $\mathcal{F}_{g}^{p}$. Analogous to FRD, the FPD is formulated as:
\begin{equation}
    \label{eq:fpd}
    \texttt{FPD}(\mathcal{R}, \mathcal{G}) = \| \mu_{g}^{p} - \mu_{r}^{p} \|_{2}^{2} + \texttt{Tr}\left(\Sigma_{g}^{p} + \Sigma_{r}^{p} - 2(\Sigma_{g}^{p}\Sigma_{r}^{p})^{\frac{1}{2}}\right)~,
\end{equation}
where $\mu_{r}^{p}$ and $\mu_{g}^{p}$ denote the mean feature embeddings of $\mathcal{F}_{r}^{p}$ and $\mathcal{F}_{g}^{p}$, and $\Sigma_{r}^{p}$ and $\Sigma_{g}^{p}$ are their corresponding covariance matrices. A smaller FPD score suggests that the generated point distributions closely resemble those of the real-world LiDAR data in the latent geometric space, capturing fine-grained structural details beyond surface-level similarities.

\vspace{0.1cm}
\noindent$\bullet$~\textit{Jensen-Shannon Divergence (JSD).}
JSD measures the similarity between the spatial occupancy distributions of real and generated LiDAR scenes from the bird's-eye-view (BEV) perspective. For each sample, we compute a 2D occupancy histogram projected onto the BEV plane, resulting in two probability distributions, denoted as $P$ (real) and $Q$ (generated). The JSD is then defined as:
\begin{equation}
    \label{eq:jsd}
    \texttt{JSD}(P \| Q) = \frac{\texttt{KL}(P \| M)}{2} + \frac{\texttt{KL}(Q \| M)}{2}~,
\end{equation}
where $M = (P + Q) / 2$ and $\texttt{KL}(\cdot \| \cdot)$ denotes the Kullback-Leibler divergence. A lower JSD value indicates that the generated BEV occupancy maps better approximate the global spatial distribution of real scenes.

\vspace{0.1cm}
\noindent$\bullet$~\textit{Maximum Mean Discrepancy (MMD).}
MMD also evaluates distributional similarity in the BEV domain, but from a kernel-based perspective. Given the same occupancy histograms $P$ and $Q$, MMD is defined as:
\begin{equation}
    \label{eq:mmd}
    \texttt{MMD}(P, Q) = \| \mathbb{E}_{P}\left[\phi(x)\right] - \mathbb{E}_{Q}\left[\phi(y)\right] \|_{\mathcal{H}}^{2}.
\end{equation}
where $\phi(\cdot)$ denotes the feature mapping to a reproducing kernel Hilbert space (RKHS). A smaller MMD value reflects a higher degree of alignment between the real and generated distributions, complementing JSD by providing a non-parametric statistical measure.

\vspace{0.1cm}
\noindent\textbf{Temporal Coherence.}
Temporal coherence measures how consistently the generated LiDAR scenes evolve over time, ensuring smooth object motion and realistic scene dynamics. Following~\cite{liang2026lidarcrafter}, we adopt two metrics:

\vspace{0.1cm}
\noindent$\bullet$~\textit{Temporal Transformation Consistency Error (TTCE).}
TTCE evaluates the temporal coherence of generated LiDAR sequences by comparing frame-to-frame transformations against ground truth. We first apply the Iterative Closest Point (ICP) algorithm~\cite{besl1992icp} to consecutive generated frames to estimate the rigid transformation $\mathbf{T}_{t}^{p} = [\mathbf{R}_{t}^{p} \mid \mathbf{t}_{t}^{p}]$, where $\mathbf{R}_{t}^{p}$ and $\mathbf{t}_{t}^{p}$ denote rotation and translation, respectively. Let $\mathbf{T}_{t}^{g} = [\mathbf{R}_{t}^{g} \mid \mathbf{t}_{t}^{g}]$ be the ground-truth transformation. TTCE is computed as:
\begin{align}
    \label{eq:ttcerot}
    \texttt{TTCErot} &= \frac{1}{T - 1} \sum_{t=1}^{T-1} \|\mathbf{R}_{t}^{p}(\mathbf{R}_{t}^{g})^{\top} - \mathbf{I}\|_{F}~, \\
    \label{eq:ttcetrans}
    \texttt{TTCEtrans} &= \frac{1}{T - 1} \sum_{t=1}^{T-1} \|\mathbf{t}_{t}^{p} - \mathbf{t}_{t}^{p}\|_{2}~,
\end{align}
where $T$ is the number of frames and $\|\cdot\|_{F}$ denotes the Frobenius norm. Lower TTCE values indicate better alignment with the ground-truth transformations, reflecting higher temporal consistency and more realistic motion in the generated LiDAR sequences.

\vspace{0.1cm}
\noindent$\bullet$~\textit{Chamfer Temporal Consistency (CTC).}
CTC measures temporal smoothness at the geometric level by computing the Chamfer Distance (CD) between consecutive frames after aligning them using ground-truth transformations. Let $\mathcal{P}_{t}$ and $\mathcal{P}_{t+1}$ be generated point clouds at frames $t$ and $t+1$. We align $\mathcal{P}_{t+1}$ to frame $t$ via:
\begin{equation}
    \mathcal{\hat{P}}_{t+1} = (\mathbf{R}_{t}^{g})^{-1}(\mathcal{P}_{t+1} - \mathbf{t}_{t}^{g})~.
\end{equation}
The Chamfer Distance between $\mathcal{P}_{t}$ and $\mathcal{\hat{P}}_{t+1}$ is:
\begin{align}
    \texttt{CD}(\mathcal{P}_{t}, \mathcal{\hat{P}}_{t+1}) =& \frac{1}{|\mathcal{P}_{t}|} \sum_{x\in\mathcal{P}_{t}} \min_{y\in\mathcal{\hat{P}}_{t+1}} \|x-y\|_{2}^{2} + \nonumber \\
    & \frac{1}{|\mathcal{\hat{P}}_{t+1}|} \sum_{y\in\mathcal{\hat{P}}_{t+1}} \min_{x\in\mathcal{P}_{t}} \|y-x\|_{2}^{2}~.
\end{align}
CTC is then averaged across all consecutive frame pairs:
\begin{equation}
    \texttt{CTC} = \frac{1}{T-1} \sum_{t=1}^{T-1} \texttt{CD}(\mathcal{P}_{t}, \mathcal{\hat{P}}_{t+1})~.
\end{equation}
Lower CTC values indicate smoother frame-to-frame transitions, reflecting stronger temporal coherence in the generated 4D LiDAR sequences.

\vspace{0.1cm}
\noindent\textbf{Downstream Utility.}
To evaluate the practical usefulness of generated LiDAR sequences for real-world perception tasks, we consider two downstream metrics: semantic segmentation performance and model calibration.

\vspace{0.1cm}
\noindent$\bullet$~\textit{Mean Intersection-over-Union (mIoU).}
mIoU is a standard evaluation metric for semantic segmentation that quantifies the overlap between predicted and ground-truth regions over all classes. It is computed as:
\begin{equation}
    \texttt{mIoU} = \frac{1}{|\mathbb{C}|} \sum_{c \in \mathbb{C}} \frac{\texttt{TP}_{c}}{\texttt{TP}_{c} + \texttt{FP}_{c} + \texttt{FN}_{c}}~, 
\end{equation}
where $\mathbb{C}$ denotes the set of semantic classes, and $\texttt{TP}_{c}$, $\texttt{FP}_{c}$, and $\texttt{FN}_{c}$ represent the number of true positives, false positives, and false negatives for class $c$, respectively. Higher mIoU values indicate more accurate segmentation and better utilization of generated LiDAR scenes in downstream perception tasks.

\vspace{0.1cm}
\noindent$\bullet$~\textit{Expected Calibration Error (ECE).}
ECE is an important metric that evaluates the calibration of a perception model, \ie, how well the predicted confidence aligns with the actual accuracy. It is defined as:
\begin{equation}
    \texttt{ECE} = \frac{1}{M} \sum_{m=1}^{M} \frac{|B_{m}|}{N} |\texttt{acc}(B_{m}) - \texttt{conf}(B_{m})|~,
\end{equation}
where $M$ is the number of confidence bins, $N$ is the total number of samples, and $|B_{m}|$ is the number of samples falling into the $m$-th bin. $\texttt{acc}(B_{m})$ and $\texttt{conf}(B_{m})$ denote the empirical accuracy and average confidence of bin $B_{m}$, respectively. Lower ECE values indicate better calibration, meaning the model’s predicted probabilities closely match the true likelihood of correctness, which is particularly important for evaluating uncertainty-aware modeling.

\section{Additional Quantitative Results}

In this section, we present additional quantitative results to further demonstrate the effectiveness of the U4D design.

\subsection{Benchmark on KITTI-360}

\begin{table}[t]
    \centering
    \caption{\textbf{Comparison of state-of-the-art LiDAR scene generation methods} on the \textit{KITTI-360}~\cite{liao2023kitti-360} dataset. Metrics marked with $\downarrow$ indicate that lower values are better. The \textbf{MMD} scores are reported in units of $10^{-4}$. The \textbf{best} and \underline{second-best} scores are highlighted in \textbf{bold} and \underline{underline}, respectively.}
    \vspace{-0.3cm}
    \label{tab:benchmark_kitti360}
    \resizebox{\linewidth}{!}{
    \begin{tabular}{r|r|cccc}
        \toprule
        \textbf{Method} & \textbf{Venue} & \textbf{FRD}~$\downarrow$ & \textbf{FPD}~$\downarrow$ & \textbf{JSD}~$\downarrow$ & \textbf{MMD}~$\downarrow$
        \\\midrule\midrule
        LiDARGen~\cite{zyrianov2022lidargen} & ECCV'22 & $579.39$ & $90.29$ & $0.07$ & $7.39$
        \\
        R2DM~\cite{nakashima2024r2dm} & ICRA'24 & \underline{$153.73$} & \underline{$6.24$} & $\mathbf{0.03}$ & \underline{$1.91$}
        \\\midrule
        \cellcolor{gen_blue!12}\textsl{U4D} & \cellcolor{gen_blue!12}\textbf{Ours} & \cellcolor{gen_blue!12}$\mathbf{142.53}$ & \cellcolor{gen_blue!12}$\mathbf{5.94}$ & \cellcolor{gen_blue!12}$\mathbf{0.03}$ & \cellcolor{gen_blue!12}$\mathbf{1.84}$
        \\\bottomrule
    \end{tabular}}
    \vspace{-0.25cm}
\end{table}

To further evaluate the effectiveness and generalization ability of U4D, we conduct additional benchmarking experiments on the widely used KITTI-360 dataset~\cite{liao2023kitti-360}. Following the same training configuration as SemanticKITTI~\cite{behley2019semantickitti}, we train and evaluate our model under identical settings to ensure a fair comparison. As reported in \cref{tab:benchmark_kitti360}, U4D consistently outperforms existing LiDAR generation methods. In particular, our method achieves lower distribution discrepancies and improved geometric fidelity, indicating that the proposed spatio-temporal modeling effectively captures both structural and temporal characteristics of LiDAR data.

\subsection{Uncertainty-Region Selection}

\begin{table}[t]
    \centering
    \caption{\textbf{Ablation study on uncertainty region selection} on the \textit{nuScenes}~\cite{caesar2020nuscenes} dataset. Metrics marked with $\downarrow$ indicate that lower values are better. The \textbf{MMD} scores are reported in units of $10^{-4}$. The \textbf{best} scores are highlighted in \textbf{bold}.}
    \vspace{-0.3cm}
    \label{tab:abl_region}
    \resizebox{\linewidth}{!}{
    \begin{tabular}{c|r|cccc}
        \toprule
        \textbf{\#} & \textbf{Threshold} & \textbf{FRD}~$\downarrow$ & \textbf{FPD}~$\downarrow$ & \textbf{JSD}~$\downarrow$ & \textbf{MMD}~$\downarrow$
        \\\midrule\midrule
        (1) & Score-based & $531.65$ & $21.53$ & $0.06$ & $0.69$
        \\\midrule
        (2) & Top-$10\%$ & $242.23$ & $13.53$ & $0.04$ & $0.55$
        \\
        (3) & Top-$15\%$ & $231.06$ & $13.02$ & $0.04$ & $0.51$
        \\
        \cellcolor{gen_blue!12}(4) & \cellcolor{gen_blue!12}Top-$20\%$ & \cellcolor{gen_blue!12}$\mathbf{223.96}$ & \cellcolor{gen_blue!12}$\mathbf{12.90}$ & \cellcolor{gen_blue!12}$\mathbf{0.03}$ & \cellcolor{gen_blue!12}$0.53$
        \\
        (5) & Top-$25\%$ & $227.53$ & $12.94$ & $\mathbf{0.03}$ & $\mathbf{0.50}$
        \\\bottomrule
    \end{tabular}}
    \vspace{-0.25cm}
\end{table}

To maintain a consistent number of uncertainty points across different scenes, we select the top-$K$ high-entropy points to form uncertainty regions. In \cref{tab:abl_region}, we conduct a series of ablation studies on the selection of the top-$K$ points. First, we directly select uncertainty points based on entropy scores (row~1). This strategy leads to poor performance, which we attribute to the large distribution gaps across scenes. As a result, the number of remaining points varies significantly, making the first-stage generation difficult to learn. To this end, we instead select the top-$K$ high-entropy points (rows~2-5), which ensures a consistent number of uncertain regions. As $K$ increases, the performance gradually improves, since more structural layout information is preserved to guide the full-scene generation. However, when $K$ becomes sufficiently large, the performance begins to saturate or even degrade, suggesting that introducing too many uncertainty points may reduce the effectiveness of the guidance.

\subsection{Segmentor for Uncertainty Regions}

\begin{table}[t]
    \centering
    \caption{\textbf{Ablation study on different segmentors for selecting uncertainty region} on the \textit{nuScenes}~\cite{caesar2020nuscenes} dataset. Metrics marked with $\downarrow$ indicate that lower values are better. The \textbf{MMD} scores are reported in units of $10^{-4}$. The \textbf{best} scores are highlighted in \textbf{bold}.}
    \vspace{-0.3cm}
    \label{tab:abl_segmentor}
    \resizebox{\linewidth}{!}{
    \begin{tabular}{c|r|cccc}
        \toprule
        \textbf{\#} & \textbf{Segmentor} & \textbf{FRD}~$\downarrow$ & \textbf{FPD}~$\downarrow$ & \textbf{JSD}~$\downarrow$ & \textbf{MMD}~$\downarrow$
        \\\midrule\midrule
        \cellcolor{gen_blue!12}(1) & \cellcolor{gen_blue!12}RangeNet++~\cite{milioto2019rangenet++} & \cellcolor{gen_blue!12}$\mathbf{223.96}$ & \cellcolor{gen_blue!12}$12.90$ & \cellcolor{gen_blue!12}$\mathbf{0.03}$ & \cellcolor{gen_blue!12}$0.53$
        \\
        (2) & FRNet~\cite{xu2025frnet} & $227.03$ & $\mathbf{12.31}$ & $\mathbf{0.03}$ & $\mathbf{0.52}$
        \\
        (3) & Cylinder3D~\cite{zhu2021cylinder3d} & $235.31$ & $12.48$ & $0.04$ & $\mathbf{0.52}$
        \\
        (4) & SPVCNN~\cite{tang2020spvcnn} & $232.56$ & $13.01$ & $\mathbf{0.03}$ & $0.54$
        \\\bottomrule
    \end{tabular}}
    \vspace{-0.25cm}
\end{table}

We adopt RangeNet++~\cite{milioto2019rangenet++} as the default segmentation model to estimate uncertainty maps. To evaluate the robustness of U4D to the choice of segmentation model for uncertainty region estimation, we experiment with several representative segmentors, including the range-view-based RangeNet++~\cite{milioto2019rangenet++} and FRNet~\cite{xu2025frnet}, the sparse-voxel-based Cylinder3D~\cite{zhu2021cylinder3d}, and the multi-view-fusion-based SPVCNN~\cite{tang2020spvcnn}. The results are summarized in \cref{tab:abl_segmentor}, U4D consistently produces high-quality LiDAR scenes regardless of the segmentation model used for uncertainty estimation. This observation indicates that U4D is not tied to a specific segmentation model and demonstrates strong robustness and generalization across different segmentors.

\section{Additional Qualitative Results}

\begin{figure*}
    \centering
    \includegraphics[width=\linewidth]{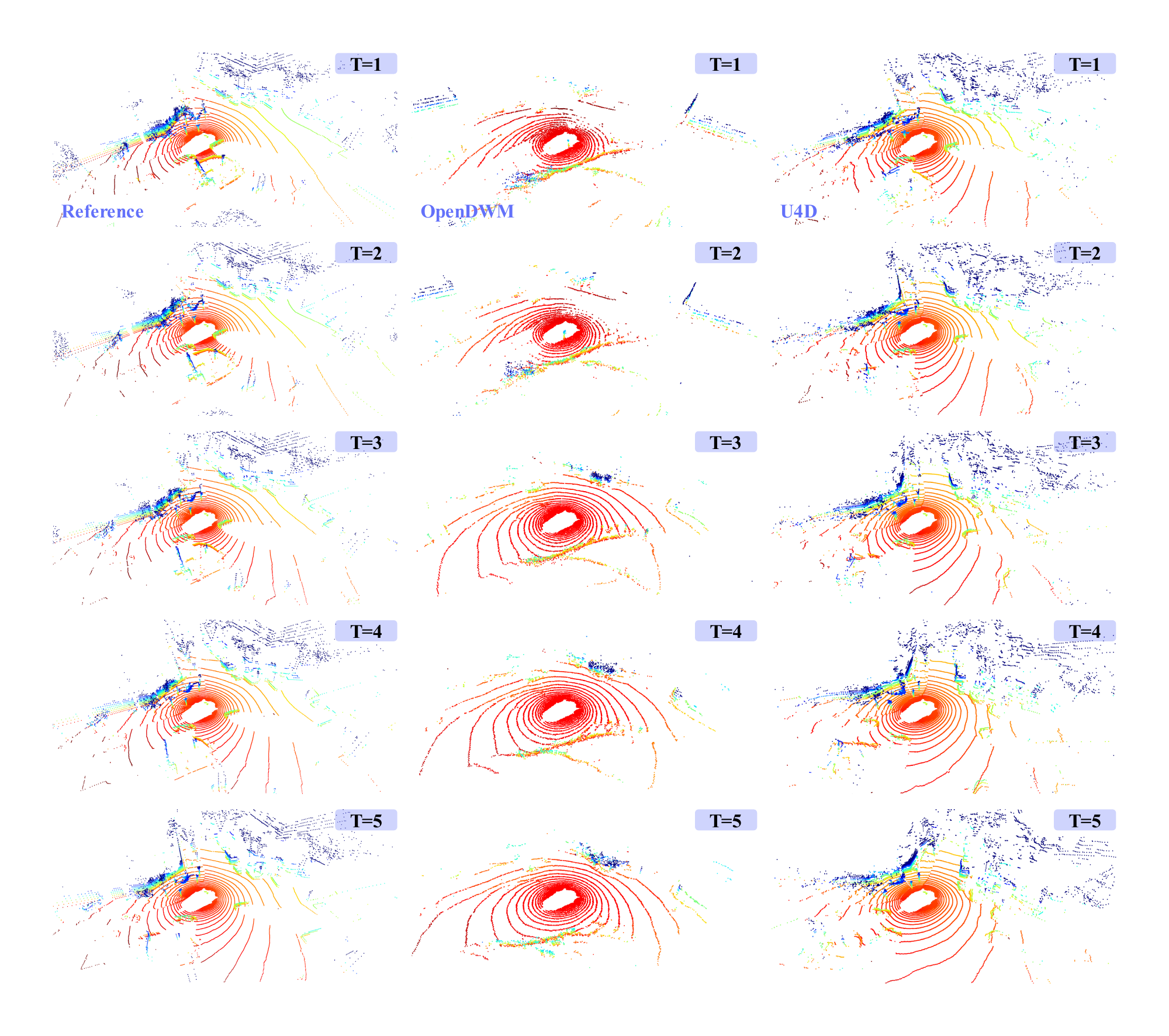}
    \caption{Qualitative results of \textbf{sequence point cloud generation} on the \textit{nuScenes} dataset~\cite{caesar2020nuscenes}. U4D preserves both geometric fidelity and temporal consistency, producing sequences most similar to the reference. It reliably reconstructs distant, sparse regions and captures dynamic objects across frames, maintaining coherent structure and motion. Frames are shown in temporal order from top to bottom. The colors are rendered based on the height information of the point cloud. Best viewed in zoom.}
    \label{fig_supp:vis_scene}
\end{figure*}

In \cref{fig_supp:vis_scene}, we present qualitative comparisons between U4D and a recent state-of-the-art LiDAR sequence generator~\cite{opendwm}, together with the corresponding reference sequences. U4D exhibits notably improved geometric fidelity and temporal coherence. It better preserves fine-grained structures that are often blurred or missing in prior methods, and in distant or low-density regions it reconstructs plausible planar surfaces with correct depth gradients. This robustness benefits from the proposed uncertainty-region modeling, which directs generation capacity toward hard-to-reconstruct areas. For dynamic objects, U4D yields smoother inter-frame transitions and more consistent object shapes and trajectories. The MoST block plays a key role here by enhancing temporal activations in intermediate layers while preserving spatial details elsewhere, enabling a more balanced spatio-temporal representation.

\begin{figure*}
    \centering
    \includegraphics[width=\linewidth]{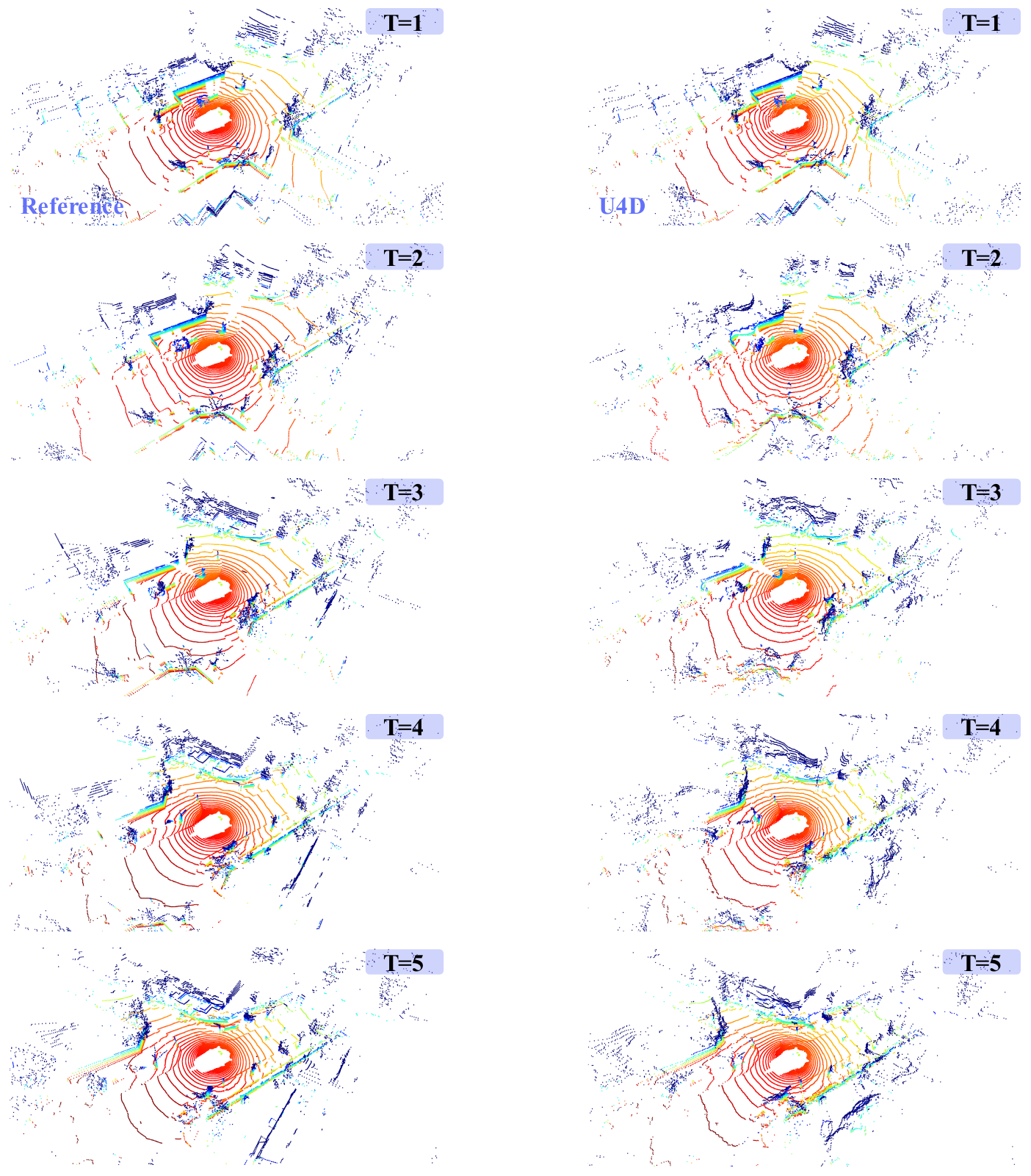}
    \caption{\textbf{Future scene prediction} on the \textit{nuScenes} dataset~\cite{caesar2020nuscenes}. Given only the first frame as input, U4D can predict plausible future LiDAR frames that maintain both geometric fidelity and temporal consistency. The model successfully captures object motion, scene evolution, and structural continuity across time. Frames are shown in temporal order from top to bottom. The colors are rendered based on the height information of the point cloud. Best viewed in zoom.}
    \label{fig_supp:vis_pred}
\end{figure*}

Beyond reconstructing observed sequences, we also explore U4D’s capability for future frame prediction. As shown in \cref{fig_supp:vis_pred}, given only the first frame, U4D can generate plausible future LiDAR observations that exhibit coherent scene evolution and realistic motion patterns. This highlights the model’s ability not only to replicate existing sequences but also to forecast future dynamics in a physically consistent manner.

\section{Broad Impact \& Limitations}

In this section, we discuss the broader impact of U4D and outline its potential limitations to provide a balanced and transparent assessment of our work.

\subsection{Broader Impact}

U4D contributes to the development of safer and more scalable autonomous driving systems by enabling high-fidelity LiDAR scene generation at both spatial and temporal levels. Its capability to synthesize realistic 4D LiDAR sequences can substantially reduce the cost of data collection and annotation, particularly for safety-critical or rare scenarios such as adverse weather, long-tail object categories, and hazardous corner cases. This can accelerate the training and benchmarking of perception models, facilitate research in uncertainty estimation, and alleviate the heavy dependency on real-world data collection, which often poses privacy, safety, and logistical challenges.

Moreover, U4D can support simulation platforms and digital-twin systems by offering controllable, diverse, and uncertainty-aware scene generation. These synthetic environments can improve reproducibility, broaden research access, and lower the entry barrier for institutions with limited resources, thereby fostering a more inclusive and equitable autonomous driving research ecosystem.

Nonetheless, as with other generative frameworks, misuse is possible. Synthetic LiDAR data could, in principle, be used to create deceptive or manipulated sensor recordings. We therefore encourage responsible and transparent use of generative models and recommend deploying proper verification and auditing mechanisms to mitigate unintended or malicious misuse.

\subsection{Potential Limitations}

Despite the strong performance of U4D, several limitations remain and point toward directions for future improvement.

\vspace{0.1cm}
\noindent\textbf{Scene Diversity and Rare Case Modeling.}
The generative capability of U4D is inherently tied to the data distribution it is trained on. While it performs robustly on common driving scenes, it may struggle to accurately reproduce extremely rare events or highly complex environments that are sparsely represented in the training set. Capturing such tail scenarios may require more diverse datasets or the integration of additional priors.

\vspace{0.1cm}
\noindent\textbf{Computational Cost.}
Although U4D adopts an efficient architecture, generating long-range temporally consistent 4D LiDAR sequences remains computationally expensive. The two-stage diffusion process requires considerable GPU resources for training, and real-time generation is still challenging for large-scale or on-vehicle deployment.

\vspace{0.1cm}
\noindent\textbf{Limited Generation Horizon.}
While U4D excels at modeling short-term temporal dynamics, its performance degrades when generating sequences longer than approximately 10 frames. This limitation mainly arises from the accumulation of stochastic errors during iterative denoising, which becomes more pronounced in the range-image space where generation is performed in a pixel-level manner. Small inconsistencies in the latent features can be amplified during decoding back into point clouds, leading to geometric drift, motion inconsistency, or structural artifacts over longer horizons. Developing more stable latent representations, stronger temporal constraints, or hierarchical generation strategies could alleviate this limitation.

\section{Public Resources Used}
\label{sec_supp:acknowledge}

In this section, we acknowledge the use of the following public resources, during the course of this work.

\subsection{Public Codebase Used}

We acknowledge the use of the following public codebase, during the course of this work:

\begin{itemize}
    \item MMEngine\footnote{\url{https://github.com/open-mmlab/mmengine}.} \dotfill Apache License 2.0

    \item MMCV\footnote{\url{https://github.com/open-mmlab/mmcv}.} \dotfill Apache License 2.0

    \item MMDetection\footnote{\url{https://github.com/open-mmlab/mmdetection}.} \dotfill Apache License 2.0

    \item MMDetection3D\footnote{\url{https://github.com/open-mmlab/mmdetection3d}.} \dotfill Apache License 2.0

    \item OpenPCDet\footnote{\url{https://github.com/open-mmlab/OpenPCDet}.} \dotfill Apache License 2.0
\end{itemize}

\subsection{Public Datasets Used}

We acknowledge the use of the following public datasets, during the course of this work:

\begin{itemize}
    \item nuScenes\footnote{\url{https://www.nuscenes.org/nuscenes}.} \dotfill CC BY-NC-SA 4.0

    \item SemanticKITTI\footnote{\url{https://semantic-kitti.org}.} \dotfill CC BY-NC-SA 4.0
\end{itemize}

\subsection{Public Implementations Used}

We acknowledge the use of the following implementations, during the course of this work:

\begin{itemize}
    \item pytorch\footnote{\url{https://github.com/pytorch/pytorch}.} \dotfill BSD License

    \item nuscenes-devkit\footnote{\url{https://github.com/nutonomy/nuscenes-devkit}.} \dotfill Apache License 2.0

    \item r2dm\footnote{\url{https://github.com/kazuto1011/r2dm}.} \dotfill MIT License

    \item lidarcrafter\footnote{\url{https://github.com/worldbench/lidarcrafter}.} \dotfill MIT License

    \item Open3D\footnote{\url{https://github.com/isl-org/Open3D}.} \dotfill MIT License

    \item torchsparse\footnote{\url{https://github.com/mit-han-lab/torchsparse}.} \dotfill MIT License

    \item LiMoE\footnote{\url{https://github.com/Xiangxu-0103/LiMoE}.} \dotfill Apache License 2.0

    \item Vista\footnote{\url{https://github.com/OpenDriveLab/Vista}.} \dotfill Apache License 2.0
\end{itemize}

\clearpage\clearpage

\clearpage\clearpage

\clearpage\clearpage
{
    \small
    \bibliographystyle{ieeenat_fullname}
    \bibliography{main}
}

\end{document}